\documentclass[times,final]{elsarticle}
\usepackage{framed,multirow}

\usepackage{amssymb}
\usepackage{latexsym}
\usepackage{bm}
\usepackage{amsmath}
\usepackage[labelfont=bf,textfont=md]{caption}
\usepackage{graphicx}
\usepackage[utf8]{inputenc}
\usepackage{tikz}
\usetikzlibrary{shapes.geometric, arrows, positioning}
\usetikzlibrary{shapes,arrows,calc}
\usepackage[ruled,vlined]{algorithm2e}
\usepackage{algorithmic}
\usepackage[utf8]{inputenc}
\usepackage[english]{babel}
\usepackage{amsthm}
\usepackage{amsmath}
\usepackage{sidecap}

\usepackage[lofdepth,lotdepth]{subfig}

\usepackage[hidelinks]{hyperref}
\hypersetup{
  colorlinks   = false, %Colours links instead of ugly boxes
  urlcolor     = blue, %Colour for external hyperlinks
  linkcolor    = green, %Colour of internal links
  citecolor   = red %Colour of citations
  }

\newcommand{\norm}[1]{\left\lVert#1\right\rVert}

\graphicspath{ {./images/} }

\usetikzlibrary{shapes.geometric, arrows}
\tikzstyle{startstop} = [rectangle, rounded corners, minimum width=3cm, minimum height=1cm,text centered, draw=black, fill=red!30]
\tikzstyle{io} = [trapezium, trapezium left angle=70, trapezium right angle=110, minimum width=3cm, minimum height=1cm, text centered, draw=black, fill=blue!30]
\tikzstyle{process} = [rectangle, minimum width=3cm, minimum height=1cm, text centered, draw=black, fill=orange!30]
\tikzstyle{decision} = [diamond, minimum width=3cm, minimum height=1cm, text centered, draw=black, fill=green!30]
\tikzstyle{arrow} = [thick,->,>=stealth]

\theoremstyle{definition}

\theoremstyle{remark}

\usepackage{url}
\usepackage{xcolor}
\definecolor{newcolor}{rgb}{.8,.349,.1}

\begin{document}

%\verso{\emph{J. Zhang, S. Zhang, G. Lin}}

\begin{frontmatter}

\title{MultiAuto-DeepONet: A Multi-resolution Autoencoder DeepONet for Nonlinear Dimension Reduction, Uncertainty Quantification and Operator Learning of Forward and Inverse Stochastic Problems\tnoteref{tnote1}}

%Rotated Multi-fidelity Gaussian process by Supervised Dimension Reduction with Iterative Rotation

% \tnotetext[tnote1]{This is an example for title footnote coding.}

\author[1]{Jiahao {Zhang}\corref{cor1}}
% \tnotetext[1]{E-mail addresses: zhan2585@purdue.edu}

\author[1]{Shiqi {Zhang}\corref{cor1}}
% \fntext[2]{E-mail addresses: zhan2296@purdue.edu}

\author[1,2]{Guang {Lin}\corref{cor2}}
%% Third author's email
% \ead{author3@author.com}
% \author[2]{Given-name4 \snm{Surname4}}

\cortext[cor1]{These authors contributed equally.}

\cortext[cor2]{Corresponding author. 
%\\ \indent\indent
% Fax: 765 494 0548; Tel: 765 494 1965; 
E-mail: guanglin@purdue.edu.}

\address[1]{Department of Mathematics, Purdue University, West Lafayette, IN 47906, USA}
\address[2]{School of Mechanical Engineering, Department of Statistics (Courtesy), Department of Earth, Atmospheric, and Planetary Sciences (Courtesy), Purdue University, West Lafayette, IN 47907, USA}

% \received{1 May 2013}
% \finalform{10 May 2013}
% \accepted{13 May 2013}
% \availableonline{15 May 2013}
% \communicated{S. Sarkar}

% MSC codes here, in the form: \MSC code \sep code
% or \MSC[2008] code \sep code (2000 is the default)
% \MSC 41A05\sep 41A10\sep 65D05\sep 65D17
% Keywords

\begin{abstract}
A new data-driven method for operator learning of stochastic differential equations(SDE) is proposed in this paper. The central goal is to solve forward and inverse stochastic problems more effectively using limited data. Deep operator network(DeepONet) has been proposed recently for operator learning. Compared to other neural networks to learn functions, it aims at the problem of learning nonlinear operators. However, it can be challenging by using the original model to learn nonlinear operators for high-dimensional stochastic problems. We propose a new multi-resolution autoencoder DeepONet model referred to as MultiAuto-DeepONet to deal with this difficulty with the aid of convolutional autoencoder. The encoder part of the network is designed to reduce the dimensionality as well as discover the hidden features of high-dimensional stochastic inputs. The decoder is designed to have a special structure, i.e. in the form of DeepONet. The first DeepONet in decoder is designed to reconstruct the input function involving randomness while the second one is used to approximate the solution of desired equations. Those two DeepONets has a common branch net and two independent trunk nets. This architecture enables us to deal with multi-resolution inputs naturally. By adding $L_1$ regularization to our network, we found the outputs from the branch net and two trunk nets all have sparse structures. This reduces the number of trainable parameters in the neural network thus making the model more efficient. Finally, we conduct several numerical experiments to illustrate the effectiveness of our proposed MultiAuto-DeepONet model with uncertainty quantification.
%DeepONet consists of a Branch Net to encode the input functions and a Trunk Net to encode the output functions. 
%especially when the dimension of physical space or stochastic operator is relatively high
\end{abstract}

\end{frontmatter}

%\linenumbers

%%%%%%%%%%%%%%%%%%%%%%%%%%%%%%%%%%%%%%%%%%%%%%%%%%%%%%%%%%%%%%%%%%%%%%%%%%%%%%%%%%%%%%%%%%%%%%%%%%%%%%%%%%%%%%%%%%%%%%
\section{Introduction}
Nowadays the fast development of computational hardware and the ability to process vast amount of data efficiently has made data-driven techniques, especially deep learning one of the most popular method to solve complex problems across many disciplines. Moreover, machine learning(ML) models can be assisted by known physical information to increase the accuracy and alleviate the amount of data required. The additional physical knowledge often appears in the form of differential equations(DEs). There has been a vast amount of work recently with different approaches such as Gaussian process(GP) \cite{sarkka2011linear, NumGP, graepel2003solving, pang2019neural, raissi2017machine, raissi2018hidden} and deep neural networks(DNNs) \cite{lagaris1998artificial, lagaris2000neural, khoo2021solving, PINN} for solving those kinds of problems. In this work, we consider a special kind of DEs, namely stochastic differential equations(SDEs) which generalized DEs by introducing randomness in coefficients or forcing terms. SDEs arise in a variety of applications naturally, for example mathematical biology, statistical mechanics, etc. The most popular approach to solve SDEs is the Monte Carlo method \cite{milstein2009solving}. However as we known, it suffers from slow rate of convergence and often requires large amount of samples to achieve certain accuracy. Another approach is built on the truncation of stochastic polynomial chaos expansion(PCE). Interested readers are referred to \cite{di2009malliavin, holden1996stochastic} for more information on this topic. Here, we want to learn stochastic operators in the setting of both the forward and inverse problems. The stochastic operators can be high-dimensional and classical numerical methods for solving this kind of problems suffer from the curse of dimensionality \cite{bellman1966dynamic} because the reliance of the carefully generated spatio-temporal grids. However, modern ML based techniques without numerical discretization is easily scalable to high dimensions.
%\cite{}[2 - 8].

In this paper we focus on DeepONet, first introduced by Lu et al. in \cite{DONet}, for stochastic operator learning. We know that the universal approximation theorem insures that neural networks(NNs) of arbitrary width or depth can approximate any continuous function of real numbers. This is the reason why DNNs become one of the most popular paradigms to solve problems relate to DEs. In their paper, the authors have designed a new deep learning framework, i.e. DeepONet to learn continuous linear or nonlinear operators. DeepONet is inspired by a similar theorem, the universal operator approximation theorem \cite{chen1995universal}. This theorem guarantees that a single hidden layer NN can approximate any nonlinear continuous operators accurately. The network inputs consist of a series of function values at fixed sensor locations, while the output sensors can be placed at any locations. The architecture of the network mainly contains two parts: one or several branch networks(stacked or unstacked DeepONet) and a trunk network take the information contained in the input sensors and the output sensors respectively. The loss function is the mean square error(MSE) between the true values and the predictions by neural network. Specially, DeepONet can be applied to learn stochastic operators. In their numerical example, the Karhunen-Loève(KL) expansion is used to deal with the random process as the input of branch network:
\begin{equation}\label{e1}
k(t; \omega) = \sum_{i=1}^N \sqrt{\lambda_i} e_i(t) \xi_i(\omega)
\end{equation}
where $\lambda_i$ and $e_i(t)$ are the $i$-th largest eigenvalue and its normalized eigenfunction of the covariance function. In this way, the input of the branch network is of the form $\sqrt{\lambda_i} [ e_i(t_1), e_i(t_2), \cdots, e_i(t_m) ]$ and the input of the trunk network is $[t, \xi_1, \xi_2, \cdots, \xi_N]$. Here, $N$ is the number of retained modes in KL expansion and $\{\xi_i\}_{i=1}^{N}$ are independent standard Gaussian random variables. Note that according to the author, the dimension of the problem is still very high because the input of the branch net is a set of $N$ eigenfunctions and the dimension for the input of the trunk net is the sum of dimensions of the physical and random space. As the dimension of the SDE problem increases, it will become unfeasible to solve the problem using original DeepONet. In our work, an encoder is proposed to conduct dimensionality reduction first before the inputs are fed into DeepONet. Then a new network called MultiAuto-DeepONet is built for nonlinear dimension reduction and stochastic operator learning. The encoder part of our model aims at discovering the hidden features of the inputs as well as reducing the dimension, while the decoder includes two DeepONets with a shared branch net. This branch net processes the inputs relate to the random process. Two different trunk nets can deal with the inputs with different resolution. For example, one trunk net for inputs in spatial domain and the other for inputs in temporal domain. Note that the sensor locations for these two trunk nets can also be different. The loss function of MultiAuto-DeepONet consists of the reconstruction error of the input functions and MSE of the SDE solution. Furthermore, we want to quantify the uncertainties associated with our quantities of interest (QOIs) coming from the stochastic process in SDEs. As we known, uncertainty quantification(UQ) is very crucial and has drawn increasing attention in machine learning methods applied to a various real life applications \cite{abdar2021review, hullermeier2019aleatoric, mukhoti2018evaluating, kabir2020spinalnet}. In solving SDE problems, high-dimensional random variables are often the sources of uncertainty. In this paper, we mainly account for the uncertainties in solving SDEs through our MultiAuto-DeepONet model. This means we show the predicted mean and variance of the model predictions. We present four different examples to illustrate the performance of our model in the numerical example section.

Our objectives of this paper:
\begin{enumerate}
    \item Nonlinear dimensionality reduction for high-dimensional stochastic inputs using MultiAuto-DeepONet.
    
    \item Learning high-dimensional stochastic operators in the setting of solving forward and inverse stochastic problems.
    
    \item Uncertainty quantification for our proposed MultiAuto-DeepONet model in stochastic operator learning problems.
\end{enumerate}

Our contributions of this paper:
\begin{enumerate}
    \item A special autoencoder with DeepONet as its decoder is constructed to perform nonlinear supervised dimensionality reduction thus discovering the nonlinear hidden features, i.e. the nonlinear representation of the high-dimensional random process. The model can effectively save computational cost in high-dimensional problems compared to PCA based DeepONet. Moreover, uncertainty quantification can be effectively carried out with much less computational cost using our MultiAuto-DeepONet model. 
    
    \item A nonlinear basis of the random process and the SDE solutions can be automatically learned. The effectiveness of this method is demonstrated by comparing to polynomial chaos expansion (PCE). The numerical experiments show our model can achieve better accuracy than PCE with the same number of base.
    %The proposed MultiAuto-DeepONet enables DeepONet to combine with autoencoder to learn the stochastic operators. Also, 

    \item Different regularizers, i.e. $L_1$, $L_2$ and etc., can be added to output layer of the decoder for different purposes. We consider the $L_1$ regularization to introduce sparsity for learned coefficients and bases. %Similarly, other regularization like the one in physics-informed neural network (PINN) can be added to reduce the SDE residue errors.
    
    \item Multi-resolution data can be handled by our proposed MultiAuto-DeepONet model naturally to solve forward and inverse stochastic problems.

    \item Applying convolutional encoder enables our model to deal with high physical dimensional problems.
\end{enumerate}

The paper is organized as follows. In Section \ref{Autoencoder}, we give a brief introduction to autoencoder. Then our MultiAuto-DeepONet model is constructed in Section \ref{deep}. In Section \ref{NR}, four numerical examples are presented to illustrate the performance of proposed model. Conclusions and future works are provided in Section \ref{Summary}.

\section{Methodology} \label{Method}
In this section, the main building blocks of our MultiAuto-DeepONet model is introduced.

\subsection{Autoencoder}\label{Autoencoder}
As we known, autoencoders were first introduced by Hinton and the PDP group \cite{rumelhart1985learning} in the 1980s. Together with Hebbian learning rules \cite{hebb2005organization, oja1982simplified}, autoencoders lay the foundation of unsupervised learning. An autoencoder is a special type of feedforward neural network where the output is set to be the same as the input and it often consists of two main parts: an encoder and an decoder. The encoder is used to process the inputs and the decoder takes the output of encoder and uses it to reconstruct the inputs. The outputs from the encoder can be understood as a latent representation of the inputs. Autoencoder can be trained by minimizing the reconstruction error which measures the differences between the original input and output from the reconstruction process of decoder. Traditional application for autoencoder includes dimensionality reduction, feature learning and etc. We apply autoencoder in our model to find the hidden features in inputs from high-dimensional stochastic process. 

\begin{figure}[h]
\centering
\includegraphics[width=0.8\textwidth]{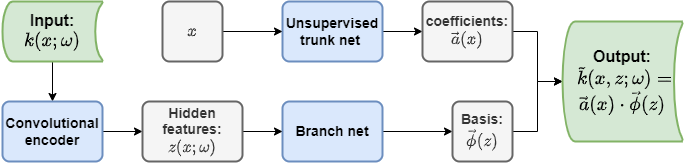}
\caption{Schematic of unsupervised part of MultiAuto-DeepONet architecture(top part of Figure \ref{MultiAuto-DeepONet}): the inputs $k(x;\omega)$ are first fed into a convolutional encoder to discover the hidden features $z(x;\omega)$. The decoder consists of one simple DeepONet. The branch net takes the hidden features as input and find a set of basis $\vec{\phi}(z)$. This set of basis is combined with coefficients $\vec{a}(x)$ from the unsupervised trunk net to compute the reconstruction $\tilde{k}(x,z;\omega)$.}
\label{auto-deeponet2}
\end{figure}

Suppose our target problem is a simple forward problem of stochastic DE, i.e. to find the solution $u(x; \omega)$. The randomness comes from the coefficient $k(x; \omega)$, where $\omega \in \Omega$ and $\Omega$ is the random space.
Figure \ref{auto-deeponet2} presents the unsupervised part of our MultiAuto-DeepONet model. We can see it is an autoencoder as stated above except that we use a convolutional encoder and the decoder part is in the form of a DeepONet. The encoder consists of several convolutional layers followed by a dense layer. The filter size, strides in convolutional layers and the activation functions in each layer depend on the specific problem and will be tuned to achieve the best performance. The hidden features $z(x; \omega)$ discovered by the convolutional encoder are fed into the branch net of the decoder. Moreover, $z(x; \omega)$ can be used to generate samples of input $k(x; \omega)$. This makes uncertainty quantification of our model much easier. Assume the number of training data is $N$, the network loss function denoted by $L_k$ for unsupervised part is exactly the reconstruction error of this autoencoder,
\begin{equation}
    L_k := \textit{MSE}_k = \frac{1}{N}\sum_{i=1}^N |k_i - \tilde{k}_i|^2
\end{equation}
where $k_i$ are training data and $\tilde{k}_i$ are network reconstruction of $k(x; \omega)$ corresponding to each $k_i$.

\begin{figure}[h]
\centering
\includegraphics[width=0.8\textwidth]{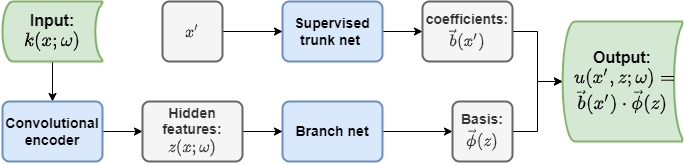}
\caption{Schematic of supervised part of MultiAuto-DeepONet architecture(bottom part of Figure \ref{MultiAuto-DeepONet}): the inputs $k(x;\omega)$ are first fed into a convolutional encoder to discover the hidden features $z(x;\omega)$. The decoder consists of one simple DeepONet. The branch net takes the hidden features as input and find a set of basis $\vec{\phi}(z)$. This set of basis is combined with coefficients $\vec{b}(x')$ from the supervised trunk net to predict solution $u(x', z;\omega)$ of the stochastic problem.}
\label{auto-deeponet3}
\end{figure}

We also present the second supervised part of our MultiAuto-DeepONet model here in Figure \ref{auto-deeponet3}. It has similar structure as unsupervised part in Figure \ref{auto-deeponet2} and they share one common branch net. The difference is that the output of supervised trunk net is used to approximate the SDE solution $u(x; \omega)$. So assume the number of training data is $M$, loss function for this supervised part denoted by $L_u$ is,
\begin{equation}
    L_u := \textit{MSE}_u = \frac{1}{M}\sum_{i=1}^M |u_i - \tilde{u}_i|^2 
\end{equation}
where $u_i$ are training data and $\tilde{u}_i$ are network prediction of $u(x; \omega)$ corresponding to each $u_i$.

\subsection{MultiAuto-DeepONet}\label{deep}
\begin{figure}[h]
\centering
\includegraphics[width=0.8\textwidth]{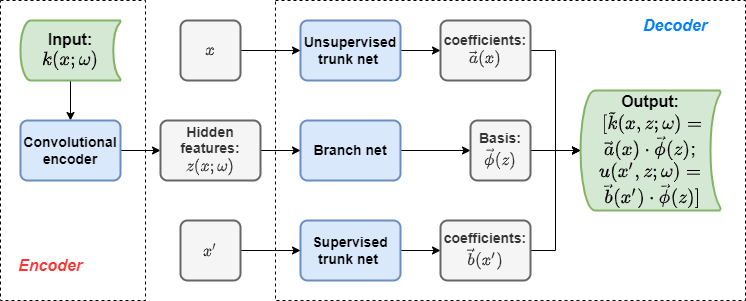}
\caption{Schematic of MultiAuto-DeepONet architecture: the inputs $k(x;\omega)$ are first fed into a convolutional encoder to discover the hidden features $z(x;\omega)$. The decoder consists of two DeepONets. The two DeepONets share a common branch net which takes the hidden features as input and find a set of basis $\vec{\phi}(z)$. This set of basis together with coefficients $\vec{a}(x)$ from the unsupervised trunk net and $\vec{b}(x')$ from the supervised trunk net are combined to compute the reconstruction $\tilde{k}(x,z;\omega)$ and predict solution $u(x',z;\omega)$.}
\label{MultiAuto-DeepONet}
\end{figure}

\begin{figure}[h]
\centering
\includegraphics[width=0.8\textwidth]{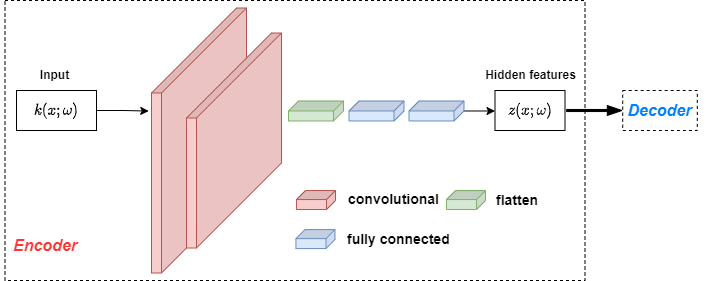}
\caption{Illustration of convolutional encoder architecture in Figure \ref{MultiAuto-DeepONet}. The input $k(x;\omega)$ is handled by the convolutional encoder to discover the hidden feature $z(x;\omega)$.}
\label{encoder}
\end{figure}

\begin{figure}[h]
\centering
\includegraphics[width=0.8\textwidth]{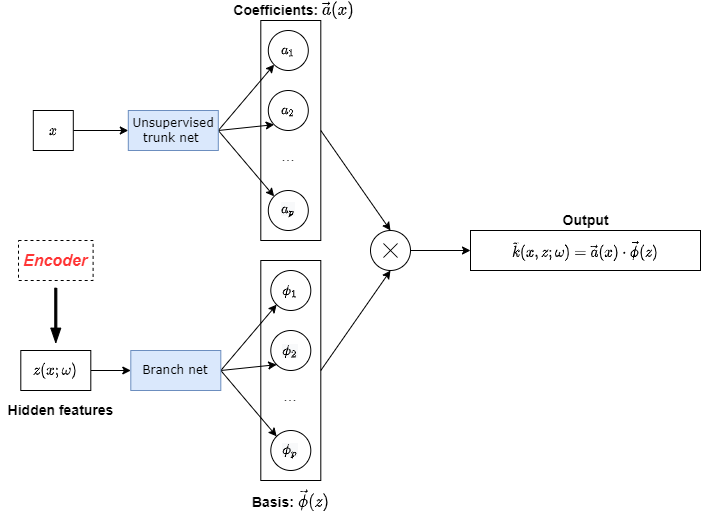}
\caption{Illustration of unsupervised part of the decoder architecture in Figure \ref{auto-deeponet2}. The input of the branch net is exactly the hidden feature $z(x;\omega)$ and this DeepONet is used to reconstruct the input $k(x;\omega)$.}
\label{DeepONet1}
\end{figure}

\begin{figure}[h]
\centering
\includegraphics[width=0.8\textwidth]{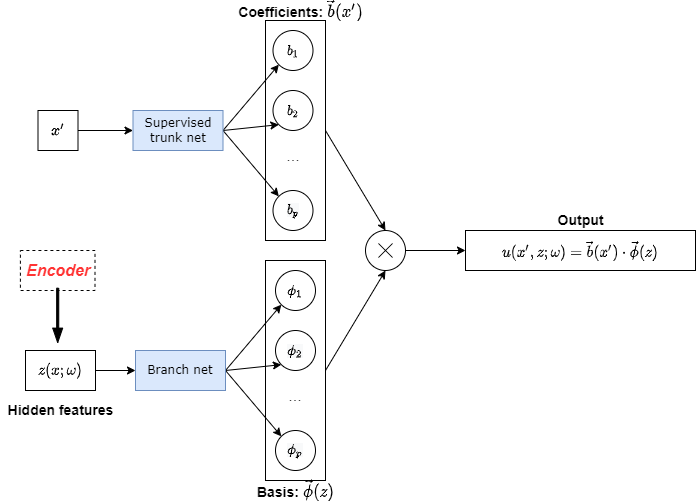}
\caption{Illustration of supervised part of the decoder architecture in Figure \ref{auto-deeponet3}. The input of the branch net is exactly the hidden feature $z(x;\omega)$ and this DeepONet is used to approximate the solution of stochastic problems.}
\label{DeepONet2}
\end{figure}

The authors in \cite{DONet} introduce DeepONet, a universal operator approximator which has shown remarkable approximation and generalization capabilities. Furthermore, they demonstrate that DeepONet can learn both explicit and implicit operators including those representing SDEs. We use DeepONet as the building block in this paper. As stated in the section of learning stochastic operators in \cite{DONet}, the input of the branch net is a random process and its dimension can be very high. As shown in Figure \ref{MultiAuto-DeepONet}, a convolutional autoencoder is used to reduce the dimensionality thus discovering the hidden features $z(x; \omega)$ from the inputs $k(x; \omega)$ first.
Then $z(x; \omega)$ is treated as the input to the common branch net of two DeepONets. The upper DeepONet in Figure \ref{MultiAuto-DeepONet} is exactly the unsupervised part in Figure \ref{auto-deeponet2} and the bottom DeepONet is the supervised part shown in Figure \ref{auto-deeponet3}. Figure \ref{encoder} presents the architecture of our convolutional encoder. The details of architectures of the two DeepONets is shown in Figures \ref{DeepONet1} and \ref{DeepONet2} respectively. Note that the unsupervised part of our MultiAuto-DeepONet model reconstructs $\tilde{k}(x; \omega)$ from the inputs and the supervised part aims at learning the solution $u(x; \omega)$. The following is the loss function of our model:
\begin{equation}\label{e2}
    L = L_k + L_u := \textit{MSE}_k + \textit{MSE}_u = \frac{1}{M}\sum_{i=1}^M |u_i - \tilde{u}_i|^2 + \frac{1}{N}\sum_{i=1}^N |k_i - \tilde{k}_i|^2
\end{equation}

Note that in MultiAuto-DeepONet model, DeepONet is used as the decoder to find a nonlinear basis for both $k(x; \omega)$ and $u(x; \omega)$. This improves the interpretation of our model. We also add $L_1$ regularization in the two DeepONets to introduce the sparse structures. Thus a relative small number of basis is required to express the solution of a complex high-dimensional problem. Other regularization can be added similarly for different purposes. The network performance and uncertainty quantification of our model will be presented in next section.

%%%%%%%%%%%%%%%%%%%%%%%%%%%%%%%%%%%%%%%%%%%%%%%%%%%%%%%%%%%%%%%%%%%%%%%%%%%%%%%%%%%%%%%%%%%%%%%%%%%%%%%%%%%%%%%%%%%%%%

\section{Numerical Results}\label{NR}
In this section, we present four numerical examples to illustrate the effectiveness of our model for nonlinear dimension reduction as well as learning stochastic operators.

\subsection{Forward problem: learning stochastic operator}
Let us first consider a pedagogical example, i.e. one dimensional stochastic ODE representing the population growth model:
\begin{equation}
    \frac{du(t; \omega)}{dt} = k(t; \omega)u(t; \omega), \hspace{3mm} t \in (0, 1] \hspace{1mm} \text{and} \hspace{1mm} \omega \in \Omega
\end{equation}
Here, $\Omega$ represents the random space and the initial condition is set to be $u(0; \omega) = 1$. As in \cite{DONet}, the coefficient $k(t; \omega)$ is modeled as the following Gaussian random process,
\begin{equation}
    k(t; \omega) \sim \mathcal{GP}(k_0(t), \text{Cov}(t_1, t_2))
\end{equation}
where $k_0(t) = 0$ is the mean funtion and the covariance function is in the squared exponential form $\text{Cov}(t_1, t_2) = \sigma^2 \text{exp}(-\frac{\norm{t_1 - t_2}^2}{2l^2})$. Here, $\sigma$ is set to be $1$ and the correlation length $l$ is in $[1, 2]$. 

The input in this problem comes from a random process and the authors in \cite{DONet} employ the Karhunen-Loeve(KL) expansion as in Equation \ref{e1} to handle it. In their example, $N$ is chosen to be $5$ which can be viewed as a truncation process. Although they did not assume to know the covariance function, the KL expansion must be known in advance. In our MultiAuto-DeepONet model, we can directly treat the observations of $k(t; \omega)$ as the inputs. The model automatically performs the nonlinear dimension reduction and discovers the hidden features first. The output of our model consists of both the solution of the equation and the reconstructed random process $k(t; \omega)$. Note that this reconstruction process of $k(t; \omega)$ can be used to generate new samples once the model is well calibrated.

\begin{figure}[h]
\centering
\subfloat[a][]{
\includegraphics[width=0.45\textwidth]{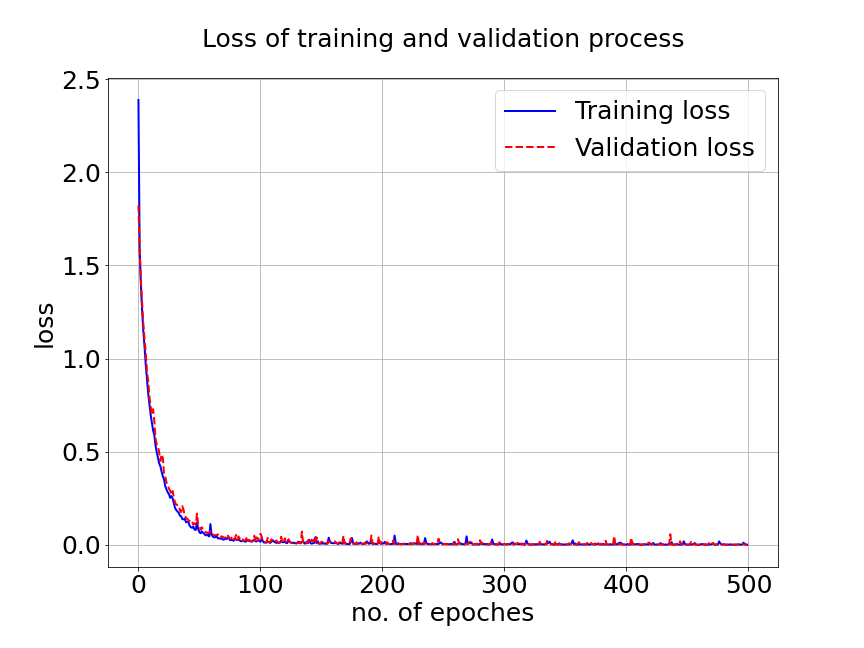}
\label{ex1_trainloss}}
\qquad 
\subfloat[b][]{
\includegraphics[width=0.45\textwidth]{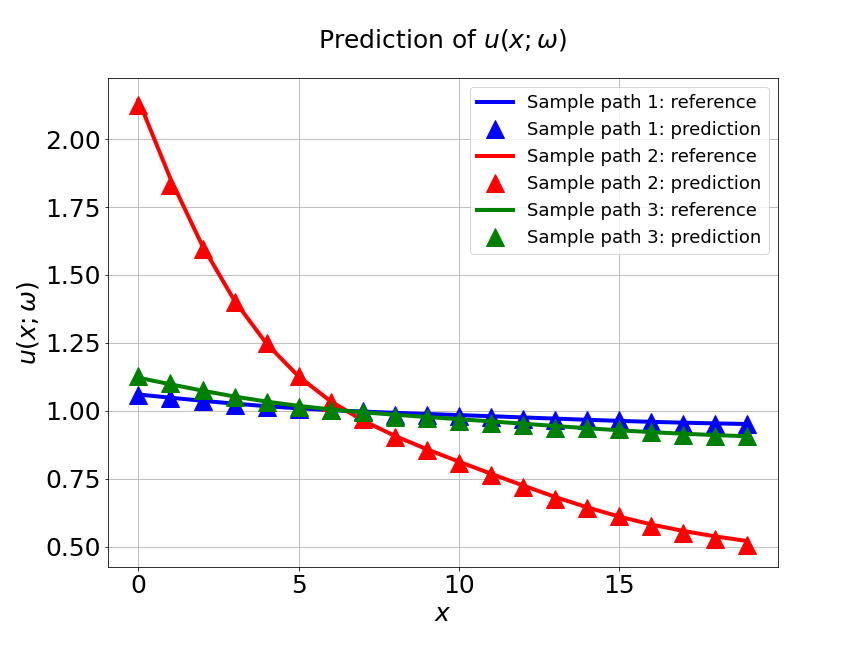}
\label{ex1_upred}}
\caption{Learning stochastic operator using MultiAuto-DeepONet: (a) The training and validation loss v.s. no. of epoches; (b) The MultiAuto-DeepONet prediction of three different sample paths. The solid lines are the reference solutions and the triangle up marker lines are the model predictions.}
\label{ex1_fig1}
\end{figure}

In our first experiment, we train MultiAuto-DeepONet model with a data set of $25,000$ different $k(t; \omega)$. The training data set consists of $1000$ samples with $25$ input sensors for each trajectories. The test MSE is approximately $0.026\%$ and the average $L^2$ relative error is around $0.52\%$. In Figure \ref{ex1_trainloss}, the blue solid line represents the training loss and the red dashed line represents the validation loss. The difference between the two losses can be ignored after hundreds of epoches. Part $(b)$ shows the predictions for three different random samples from $k(t; \omega)$. The reference solutions are solid lines with different colors and the model predictions are triangle up marker lines. We can see the predictions from MultiAuto-DeepONet model matches the reference pretty well. Table \ref{table:1} shows the MSE and average relative $l^2$ error of MultiAuto-DeepONet model for different the number of input $k$ sensors. Both errors decrease as the number of sensors increases.

\begin{table}[h!]
\centering
\begin{tabular}{||c c c||} 
 \hline
 & MSE  &  Average relative $l^2$ error
 \\ [0.5ex] 
 \hline
$n_k = 10$ &  0.00775
&  0.01122
 \\  
 \hline
 $n_k = 15$ &  0.00126
&   0.00754
 \\  
 \hline
 $n_k = 20$
 & 0.00029
 &  0.00578
 \\  
  \hline
 $n_k = 25$
 & 0.00026
 &  0.00521
 \\
   \hline
\end{tabular}
\caption{The test MSE and average $l^2$ error of MuliAuto-DeepONet model for different number of input sensors.}
\label{table:1}
\end{table}

\begin{figure}[h]
\centering
\subfloat[a][]{
\includegraphics[width=0.45\textwidth]{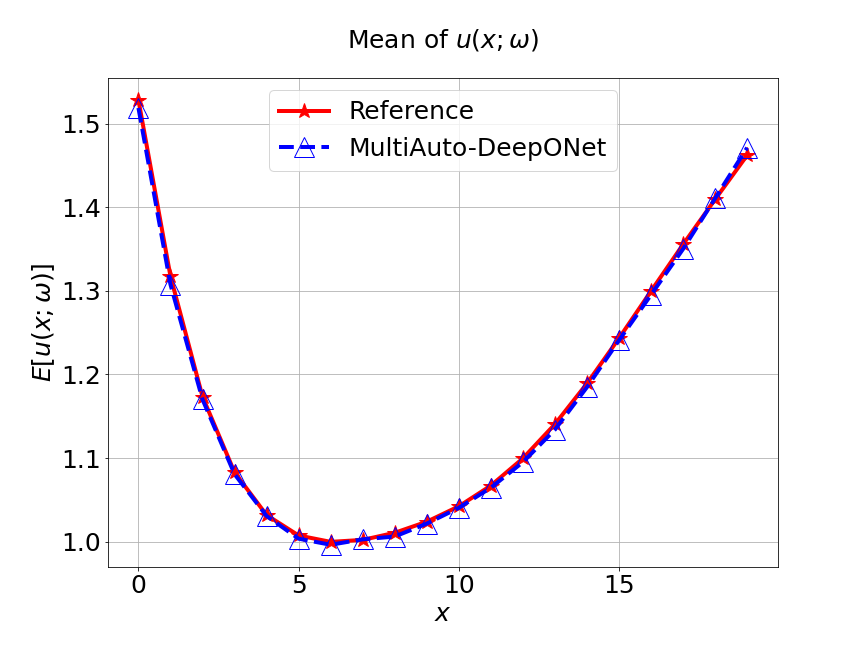}
\label{ex1_umean}}
\qquad 
\subfloat[b][]{
\includegraphics[width=0.45\textwidth]{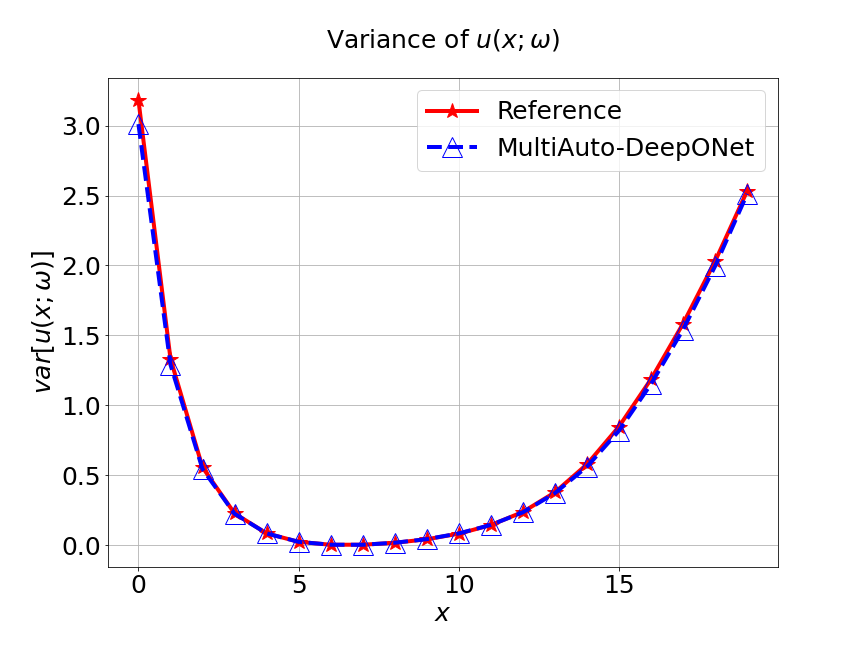}
\label{ex1_uvar}}
\qquad 
\subfloat[c][]{
\includegraphics[width=0.45\textwidth]{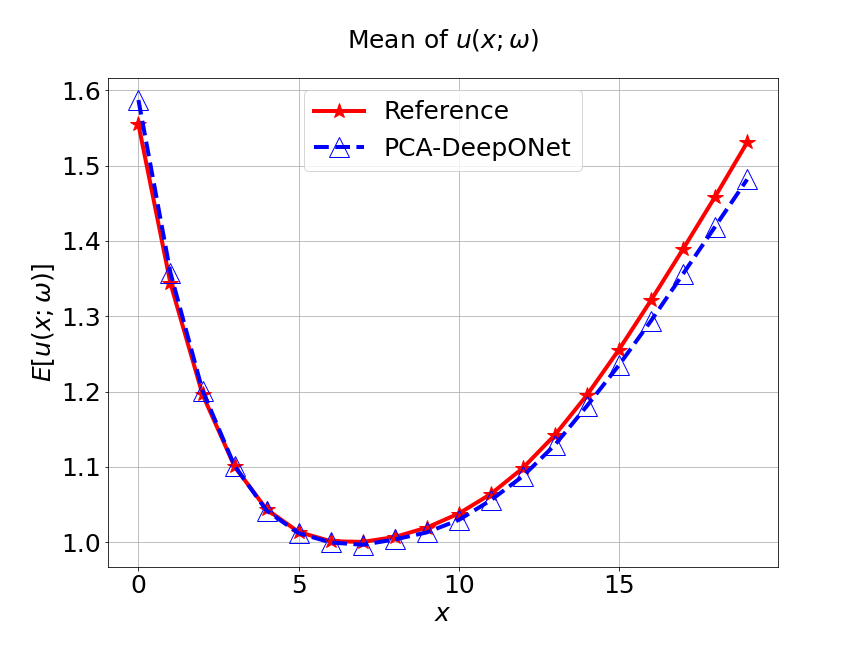}
\label{ex1_umeanPCA}}
\qquad
\subfloat[d][]{
\includegraphics[width=0.45\textwidth]{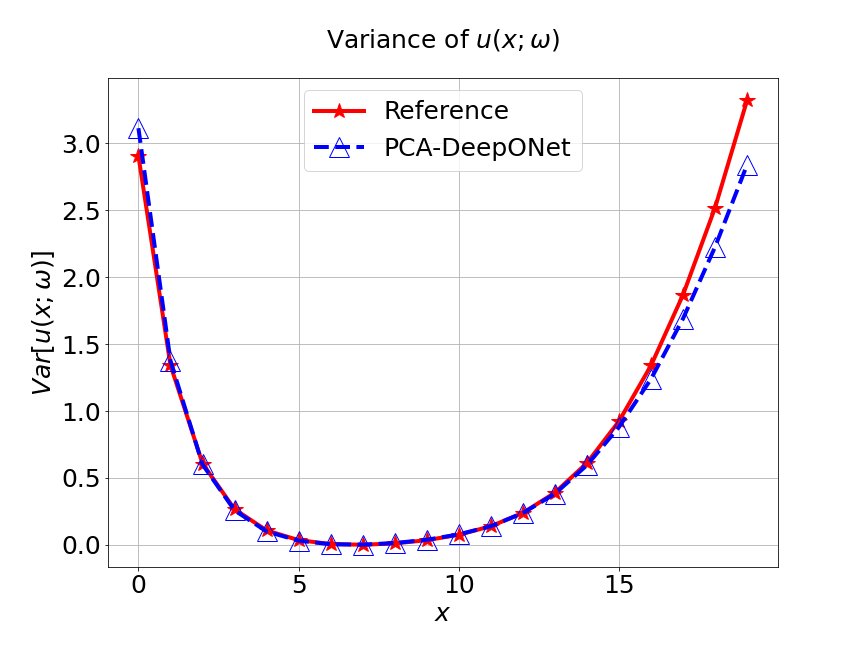}
\label{ex1_uvarPCA}}
\qquad 
\subfloat[e][]{
\includegraphics[width=0.45\textwidth]{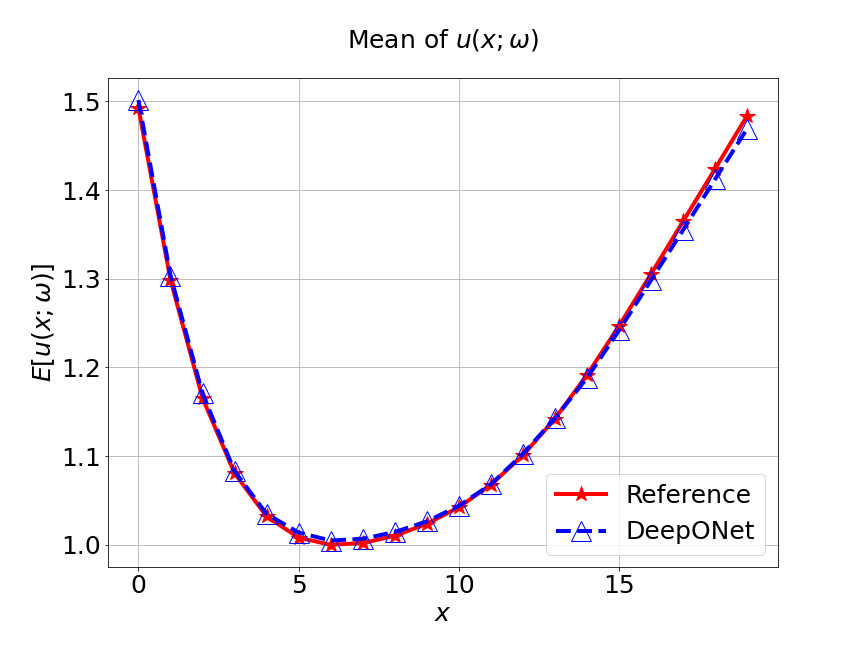}
\label{ex1_umean_DOnet}}
\qquad 
\subfloat[f][]{
\includegraphics[width=0.45\textwidth]{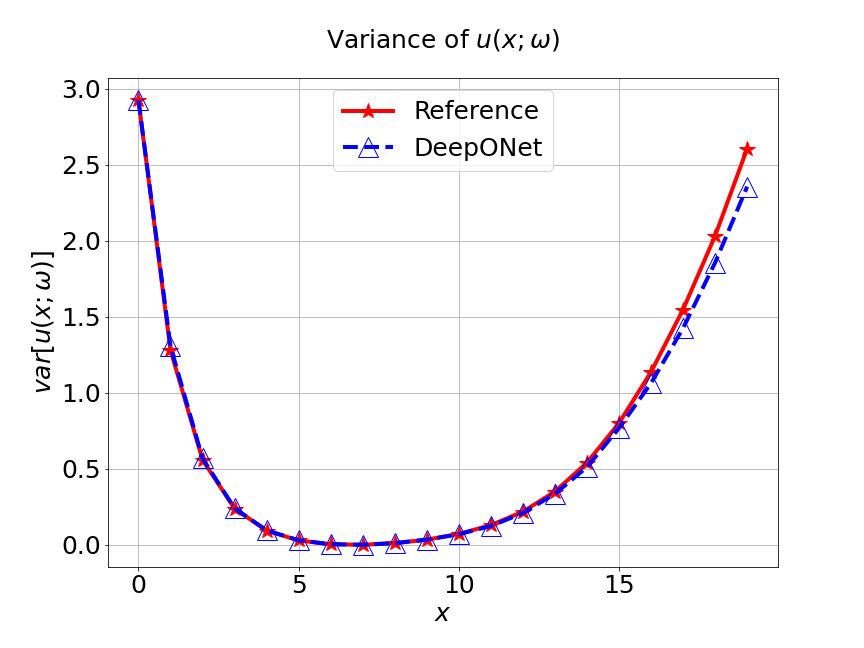}
\label{ex1_uvar_DOnet}}

\caption{Forward problem: the predicted mean and variance of different models.
(a) The predicted mean(blue triangle up line) of $u(x;\omega)$ from MultiAuto-DeepONet model and the reference solution mean(red star line); (b) The predicted variance(blue triangle up line) of $u(x;\omega)$ from MultiAuto-DeepONet model and reference solution variance(red star line); (c) The predicted mean(blue triangle up line) of $u(x;\omega)$ from PCA based DeepONet model and the reference solution mean(red star line); (d) The predicted variance(blue triangle up line) of $u(x;\omega)$ from PCA based DeepONet model and reference solution variance(red star line);(e) The predicted mean(blue triangle up line) of $u(x;\omega)$ from DeepONet model and the reference solution mean(red star line); (f) The predicted variance(blue triangle up line) of $u(x;\omega)$ from DeepONet model and reference solution variance(red star line).}
\label{ex1_meanvar}
\end{figure}

Next, we conduct a experiment to compare our model with the other two different models. The first one is the original DeepONet model. The second is PCA based DeepONet model which we call it PCA-DeepONet. This model performs PCA to the inputs from $k(t; \omega)$ instead of reducing the spatial dimension using autoencoder. Figure \ref{ex1_meanvar} presents the predicted mean and variance of the solution $u$ for the reference and different methods. Table \ref{table:2} lists the relative $l^2$ error of the predicted mean and variance when the number of input sensors is $20$. From Figure \ref{ex1_meanvar} and Table \ref{table:2}, we can see that our MultiAuto-DeepONet model achieves the best accuracy for predicted mean and variance among all three models when the training data set is relatively small.

\begin{table}[h!]
\centering
\begin{tabular}{||c c c c ||} 
 \hline
 & DeepONet  &  PCA based DeepONet &  MultiAuto-DeepONet 
 \\ [0.5ex] 
  \hline
 Mean error & 0.00530
 &  0.01686
 & \textbf{0.00459}
     \\
 \hline
Variance error & 0.0665
 & 0.0918
  &  \textbf{0.0273}
 \\  
 \hline
\end{tabular}
\caption{Comparison of relative $l^2$ errors of predicted mean and variance for different models of learning a stochastic operator.}
\label{table:2}
\end{table}

\begin{figure}[h]
\centering
\subfloat[a][]{
\includegraphics[width=0.45\textwidth]{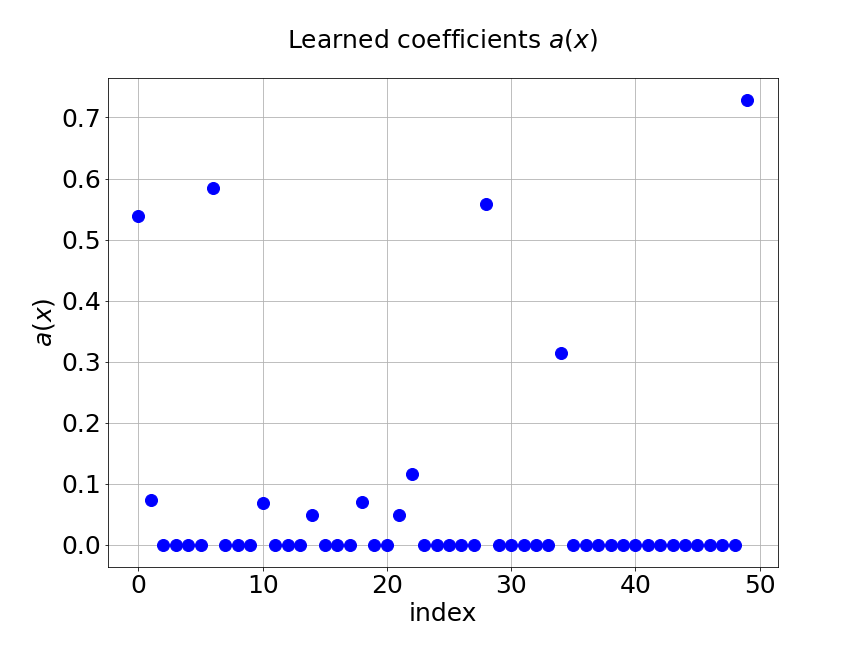}
\label{ex1_coek}}
\qquad 
\subfloat[b][]{
\includegraphics[width=0.45\textwidth]{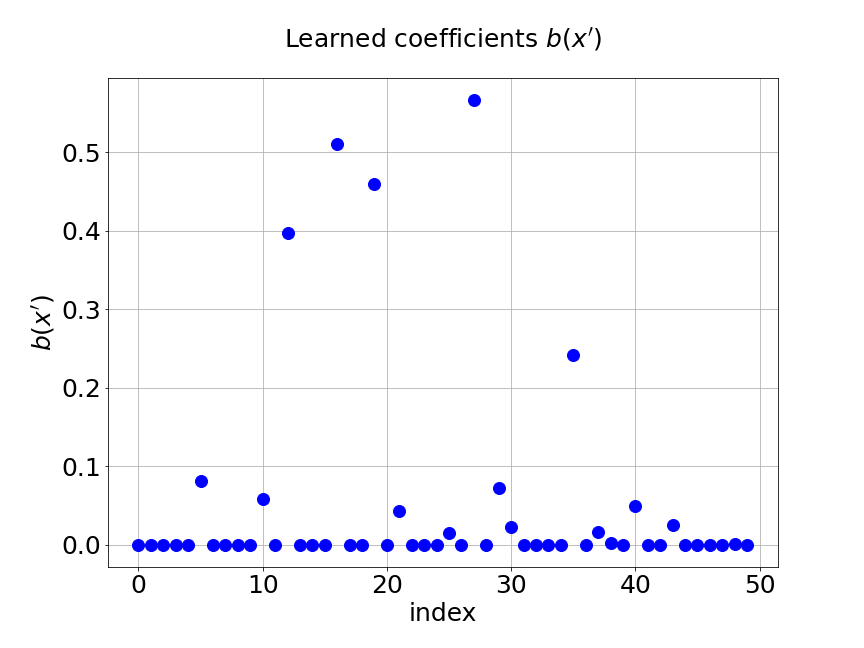}
\label{ex1_coeu}}
\qquad 
\subfloat[c][]{
\includegraphics[width=0.45\textwidth]{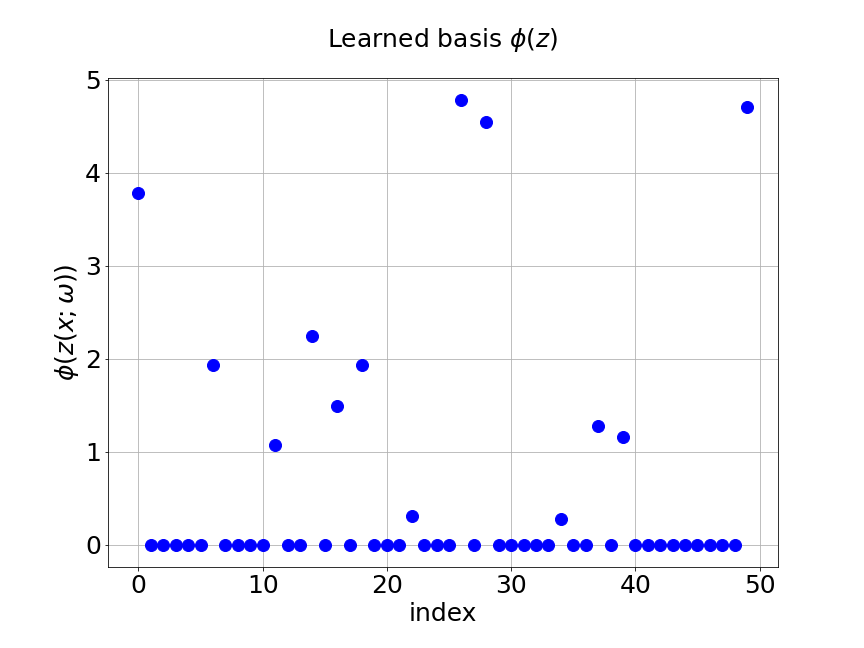}
\label{ex1_phi}}
\caption{Sparse coefficients and basis learned from MultiAuto-DeepONet for one chosen sample: (a) The coefficients $\vec{a}(x)$ from the unsupervised trunk net; (b) The coefficients $\vec{b}(x')$ from the supervised trunk net; (c) The basis $\vec{\phi}(z)$ from the branch net.}
\label{ex1_sparse}
\end{figure}

As we have mentioned, we add $L_1$ regularization in the two trunk nets of MultiAuto-DeepONet model in order to introduce the sparsity. Figure \ref{ex1_sparse} shows the coefficients and basis learned from our model for one specific samples at the corresponding output sensor locations. Part $(a)$ shows the learned coefficients from the unsupervised trunk net and part $(b)$ is the learned coefficients from the supervised trunk net. The learned basis from the branch net is presented in part $(c)$. The $x$ axes in the figures are the output sensor locations and the $y$ axes are corresponding values. We can see that only few numbers of them are not zero. This reduces the number of parameters in the network thus model training and calibration become much easier and faster.

\begin{figure}[h]
\centering
\subfloat[a][]{
\includegraphics[width=0.45\textwidth]{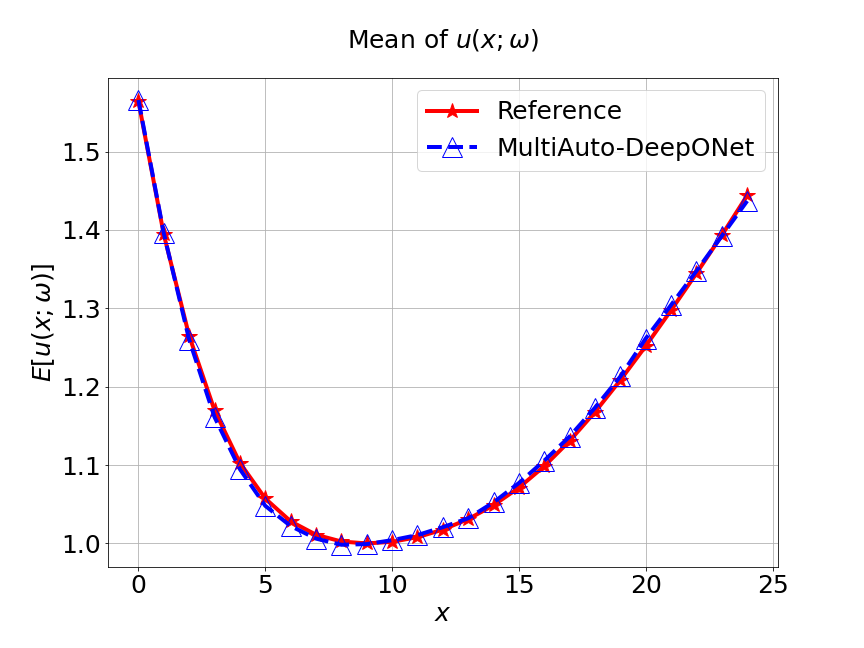}
\label{ex1_umean_gen}}
\qquad 
\subfloat[b][]{
\includegraphics[width=0.45\textwidth]{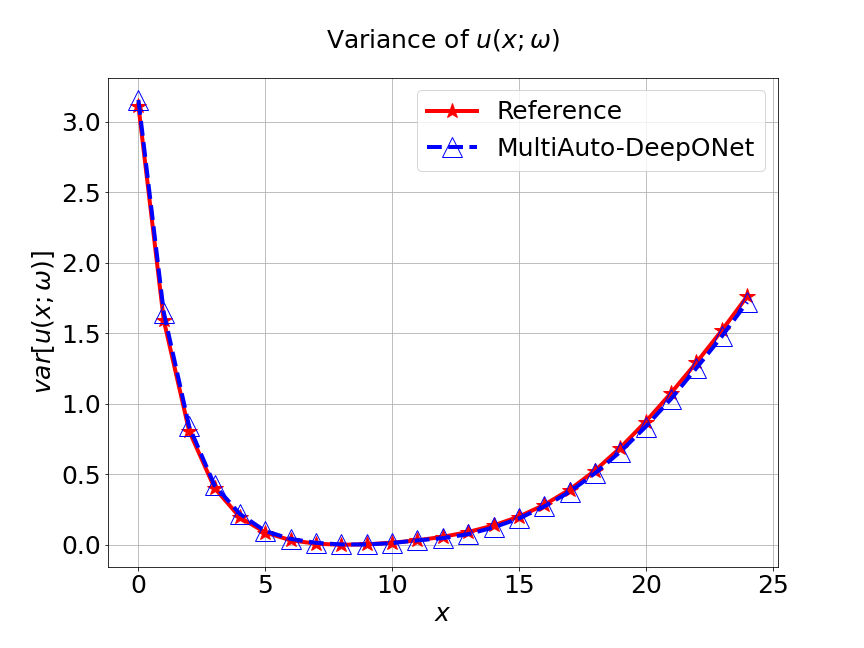}
\label{ex1_uvar_gen}}
\caption{The mean and variance of $u(x;\omega)$ from generated samples of MultiAuto-DeepONet. The method used to estimate the probability density function of $z$ is the kernel density estimation.}
\label{ex1_ugen}
\end{figure}

One major difference between our model and original DeepONet model is that we can generate samples of $k(x;\omega)$ and $u(x;\omega)$ simultaneously after the network is trained. This is achieved by using the hidden features $z(x;\omega)$ learned from the MultiAuto-DeepONet model and estimate the probability density function of $z$ to construct a generator. The method we apply in this paper is the kernel density estimation(KDE) and $3000$ samples of $u(x;\omega)$ are generated. We present the mean and variance of those samples in Figure \ref{ex1_ugen}. We can see the accuracy of the mean and variance of generated samples are relatively good considering the size of training data set.

\subsection{Inverse problem: learning forcing term of
stochastic operator}
In this example, we consider a stochastic inverse problem, i.e. the following one-dimensional stochastic Poisson equation:
\begin{equation}
    -\frac{d^2}{dx^2}u = f(x;\omega)
\end{equation}
with homogeneous boundary conditions,
\begin{equation}
    u(-1;\omega) = u(1;\omega) = 0
\end{equation}
Here, $x \in [-1, 1]$ and $\omega \in \Omega$ which is the random space and the randomness comes from the forcing term $f(x;\omega)$. We model $f(x;\omega)$ as a Gaussian process,
\begin{equation}
    f(x;\omega) \sim \mathcal{GP}(f_0(x), \text{Cov}(x_1, x_2))
\end{equation}
with mean function $f_0(x) = sin(\pi x)$ and covariance function $\text{Cov}(x_1, x_2) = \sigma^2 \text{exp}(-\frac{\norm{x_1 - x_2}^2}{2l^2})$. The standard deviation $\sigma$ is chosen to be $1$ and the correlation length $l = 1.5$.

\begin{figure}[h]
\centering
\subfloat[a][]{
\includegraphics[width=0.45\textwidth]{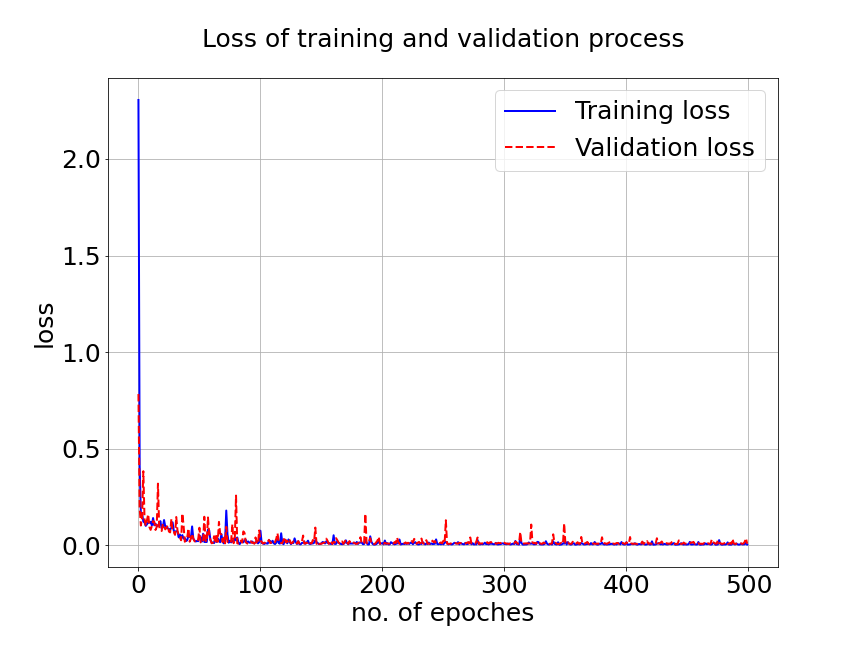}
\label{ex2_trainloss}}
\qquad 
\subfloat[b][]{
\includegraphics[width=0.45\textwidth]{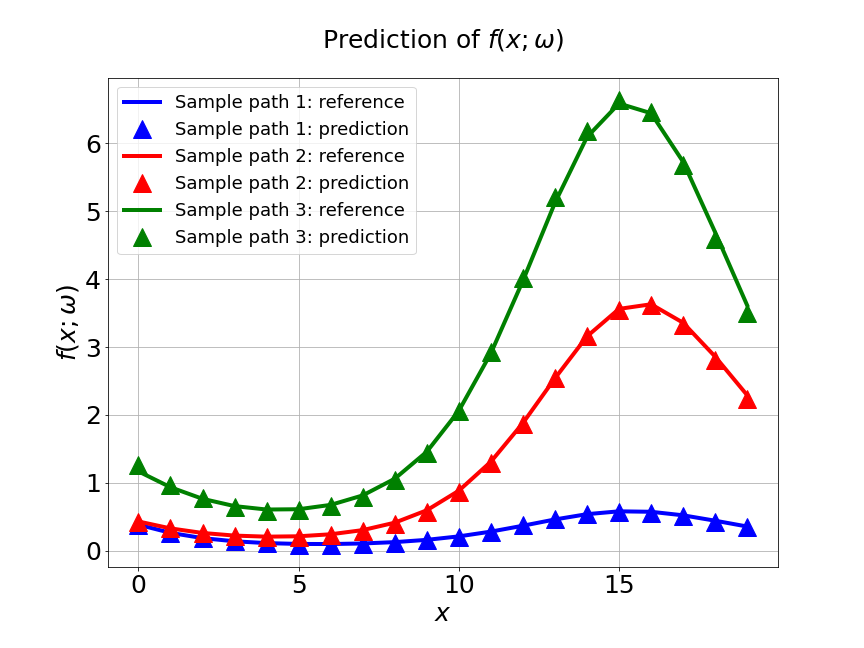}
\label{ex2_upred}}
\caption{Inverse problem: (a) The training and validation loss v.s. no. of epoches; (b) The MultiAuto-DeepONet prediction of three different sample paths. The solid lines are the reference solutions and the triangle up marker lines are the model predictions.}
\label{ex2_fig1}
\end{figure}

First, we train the model with a data set of $40,000$ different $u(x; \omega)$. The number of sample trajectories is $1000$ and the number of input $u$ sensors is $40$. The test MSE is approximately $0.12\%$ and the average $L^2$ relative error is around $0.016$. In Figure \ref{ex2_trainloss} , the blue solid line represents the training loss and the red dashed line represents the validation loss. Part $(b)$ shows the predictions for three different random samples from $f(x; \omega)$. We can see the predictions from MultiAuto-DeepONet model is accurate. Table \ref{table:3} shows the MSE and average relative $l^2$ errors of MultiAuto-DeepONet model for different number of input $u(x; \omega)$ sensors. We see they both decrease as the number of input $u$ sensors increases.

\begin{table}[h!]
\centering
\begin{tabular}{||c c c||} 
 \hline
 & MSE  &  Average relative $l^2$ error
 \\ [0.5ex] 
  \hline
$n_u = 10$  & 0.0324 &  0.0632 \\
 \hline
$n_u = 20$ &  0.0124
&  0.0464
 \\  
 \hline
 $n_u = 30$ &  0.0082
&  0.0232
 \\  
 \hline
 $n_u = 40$
 & 0.0061
 &  0.0162
 \\  
 \hline
\end{tabular}
\caption{The test MSE and average $l^2$ error of MuliAuto-DeepONet model for different number of input sensors.}
\label{table:3}
\end{table}

\begin{figure}[h]
\centering
\subfloat[a][]{
\includegraphics[width=0.45\textwidth]{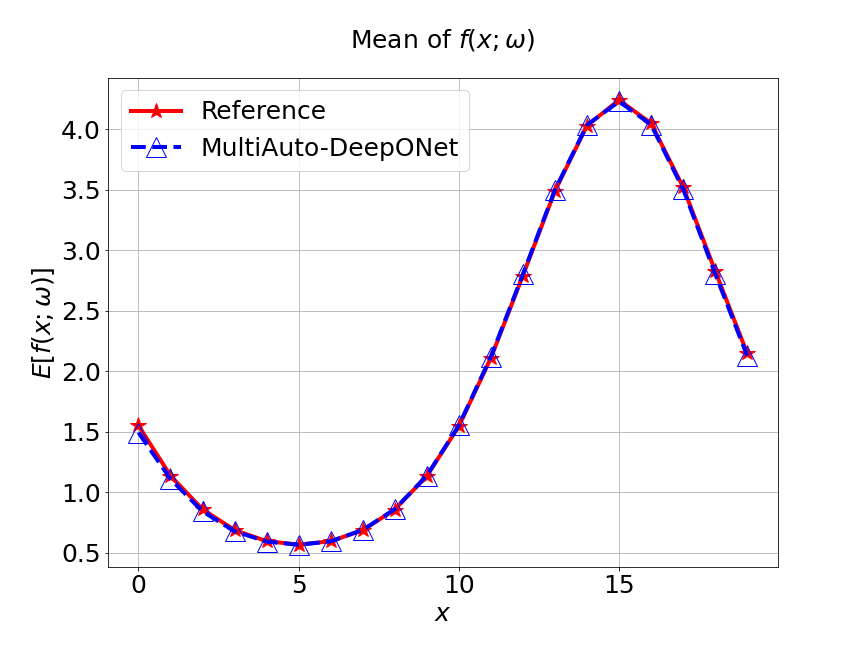}
\label{ex2_kmean}}
\qquad 
\subfloat[b][]{
\includegraphics[width=0.45\textwidth]{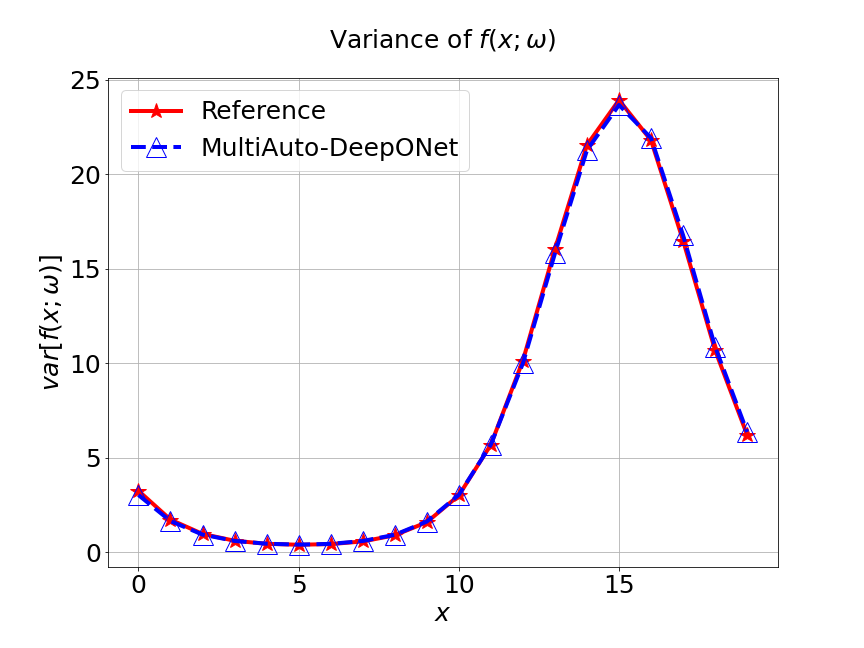}
\label{ex2_kvar}}
\qquad 
\subfloat[c][]{
\includegraphics[width=0.45\textwidth]{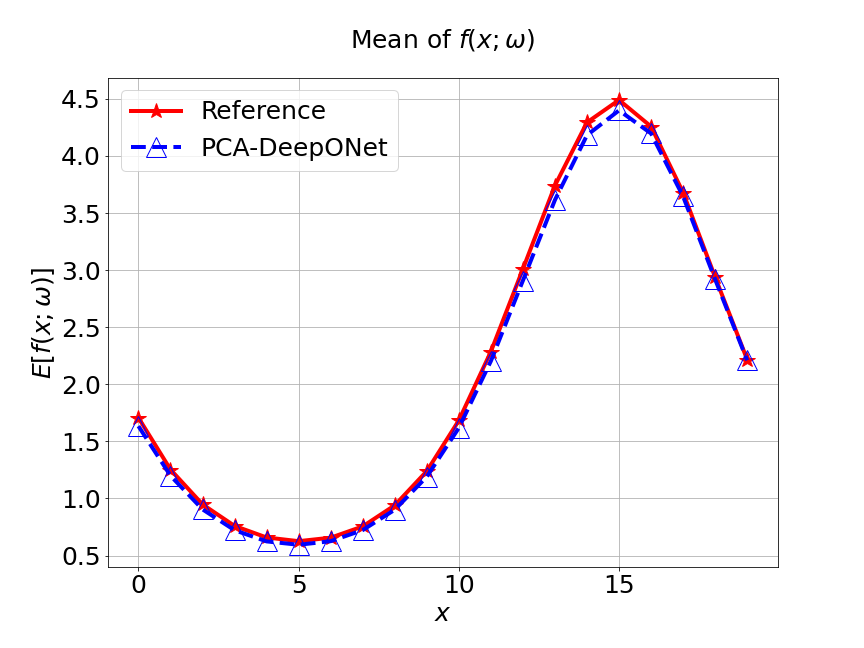}
\label{ex2_kmean_PCA}}
\qquad
\subfloat[d][]{
\includegraphics[width=0.45\textwidth]{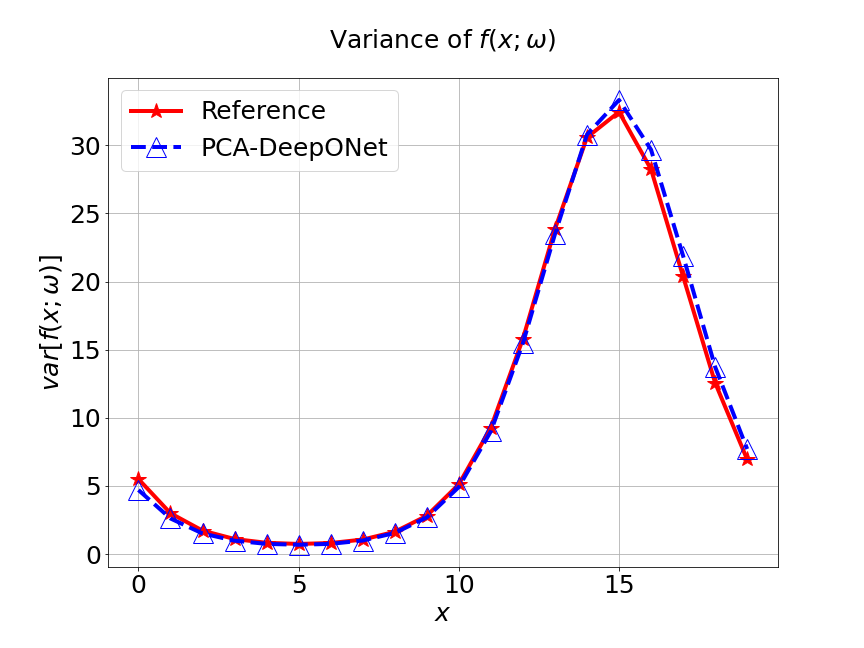}
\label{ex2_kvar_PCA}}
% \qquad 
% \subfloat[e][]{
% \includegraphics[width=0.45\textwidth]{ex2_kmean_DOnet.png}
% \label{ex2_kmean_DOnet}}
% \qquad 
% \subfloat[f][]{
% \includegraphics[width=0.45\textwidth]{ex2_kvar_DOnet.png}
% \label{ex2_kvar_DOnet}}

\caption{Inverse problem: the predicted mean and variance of different models.
(a) The predicted mean(blue triangle up line) of $f(x;\omega)$ from MultiAuto-DeepONet model and the reference solution mean(red star line); (b) The predicted variance(blue triangle up line) of $f(x;\omega)$ from MultiAuto-DeepONet model and reference solution variance(red star line); (c) The predicted mean(blue triangle up line) of $f(x;\omega)$ from PCA based DeepONet model and the reference solution mean(red star line); (d) The predicted variance(blue triangle up line) of $f(x;\omega)$ from PCA based DeepONet model and reference solution variance(red star line).
% ;(e) The predicted mean(blue triangle up line) of $f(x;\omega)$ from DeepONet model and the reference solution mean(red star line); (f) The predicted variance(blue triangle up line) of $f(x;\omega)$ from DeepONet model and reference solution variance(red star line)
}
\label{ex2_meanvar}
\end{figure}

Now, a comparison between our model and PCA based DeepONet model is performed. Figure \ref{ex2_meanvar} presents the predicted mean and variance of unknown forcing term $f$ and the reference of different methods for the inverse stochastic Poisson equation. Table \ref{table:4} lists the relative $l^2$ error of the predicted mean and variance when the number of input sensor is $40$. We can see that our MultiAuto-DeepONet model achieves the best performance for predicted mean and variance.

\begin{table}[h!]
\centering
\begin{tabular}{||c c c ||} 
 \hline
 &  PCA based DeepONet &  MultiAuto-DeepONet 
 \\ [0.5ex] 
  \hline
 Mean error &  0.0257
 & \textbf{0.0049
}
     \\
 \hline
Variance error & 0.0436
  &  \textbf{0.0136}
 \\  
 \hline
\end{tabular}
\caption{Comparison of relative $l^2$ errors of predicted mean and variance for different models of inverse problem.}
\label{table:4}
\end{table}

\begin{figure}[h]
\centering
\subfloat[a][]{
\includegraphics[width=0.45\textwidth]{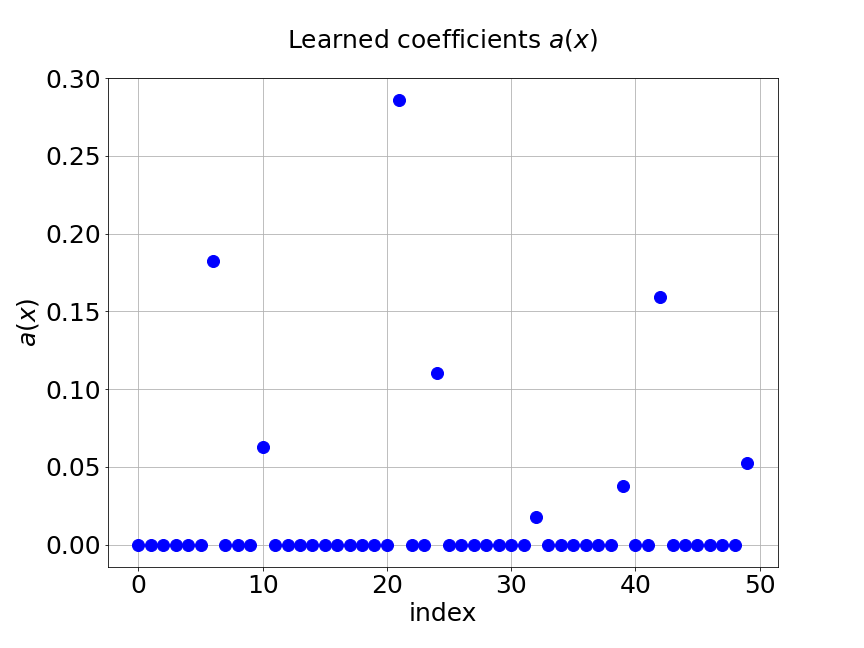}
\label{ex2_coek}}
\qquad 
\subfloat[b][]{
\includegraphics[width=0.45\textwidth]{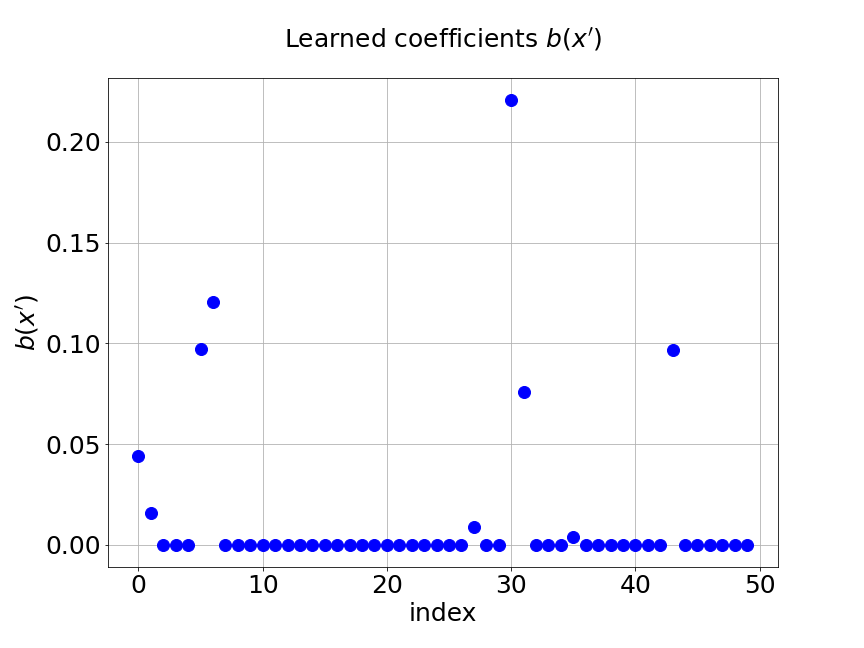}
\label{ex2_coeu}}
\qquad 
\subfloat[c][]{
\includegraphics[width=0.45\textwidth]{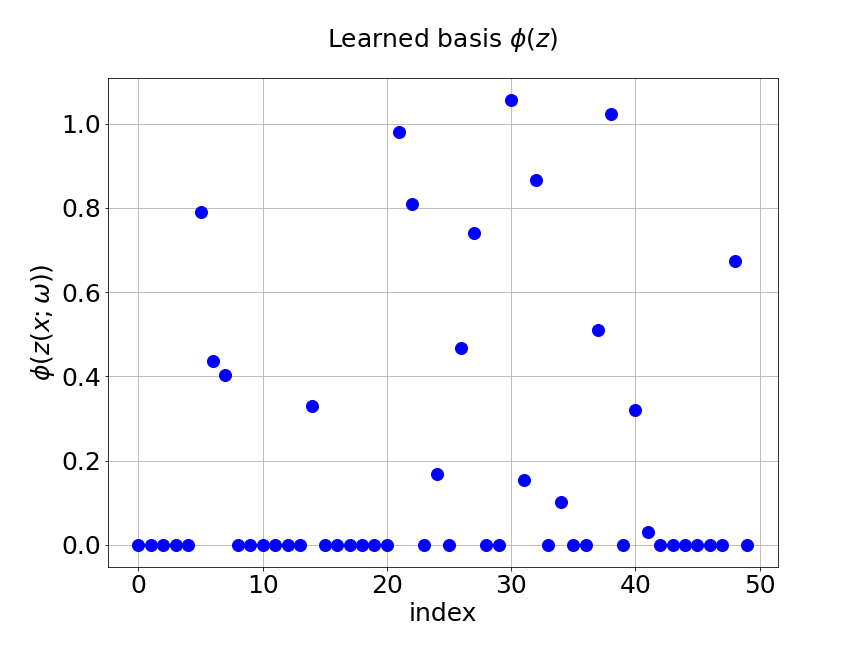}
\label{ex2_phi}}
\caption{Sparse coefficients and basis learned from MultiAuto-DeepONet: (a) The coefficients $\vec{a}(x)$ from the unsupervised trunk net; (b) The coefficients $\vec{b}(x')$ from the supervised trunk net; (c) The basis $\vec{\phi}(z)$ from the branch net.}
\label{ex2_sparse}
\end{figure}

Figure \ref{ex2_sparse} shows the coefficients learned from our model for one specific sample of $u$ at the corresponding output sensor locations. Part $(a)$ shows the learned coefficients from the unsupervised trunk net and part $(b)$ is the learned coefficients from the supervised trunk net. The learned basis from the branch net is presented in part $(c)$. The $x$ axes in the figures are the output sensor locations and the $y$ axes are corresponding values. We can see that most of those value are zero in this problem. This results from the $L_1$ regularization in the trunk nets.

\subsection{Forward problem: two dimensional Poisson Equation}
In this example, we consider the forward problem of a two-dimensional stochastic Poisson equation:
\begin{equation}
    -\frac{d^2}{dx^2}u -\frac{d^2}{dy^2}u= f(x, y;\omega), \hspace{3mm} x, y \in [-1, 1] \hspace{1mm} \text{and} \hspace{1mm} \omega \in \Omega
\end{equation}
with homogeneous boundary conditions,
\begin{equation}
    u(x, -1;\omega) = u(x, 1;\omega) = u(-1, y;\omega) = u(1, y;\omega) = 0
\end{equation}
Here, $\Omega$ is the random space and the randomness comes from the forcing term $f(\vec{t};\omega)$ ($\vec{t} = (x, y)$) as in the last example except that the spatial dimension is $2$. Again as the previous example, $f(\vec{t};\omega)$ is modelled as a Gaussian process:
\begin{equation}
    f(\vec{t};\omega) \sim \mathcal{GP}(f_0(\vec{t}), \text{Cov}(\vec{t}_1, \vec{t}_2))
\end{equation}
with mean function $f_0(\vec{t})=20sin(\pi (x+y))$ and covariance function $\text{Cov}(\vec{t}_1, \vec{t}_2)=\sigma^2 \text{exp}(-\frac{\norm{\vec{t}_1 - \vec{t}_2}^2}{2l^2})$. The hyper-parameters $\sigma$ and $l$ are set to be $0.5$ and $1.5$ respectively.

In this two dimensional problem, the latent dimension of the encoder is chosen to be $15$. There are two convolutional layers and two dense layers with \textit{relu} activation function. For convolution layers, the stride length is $1$ and the filter size is $5$. We add $L_1$ regularization in the two trunk nets of the decoder to introduce the spasity. This step is crucial because the spatial dimension is two in this problem. The Adam optimizer is applied to train the network with batch size $256$. For our training data set, the number of sample trajectories of $f(x,y;\omega)$ is $2000$ and the number of input $f(x,y;\omega)$ sensors is $400$, i.e. $20 \times 20$ grid. 

\begin{figure}[h]
\centering
\subfloat[a][]{
\includegraphics[width=0.45\textwidth]{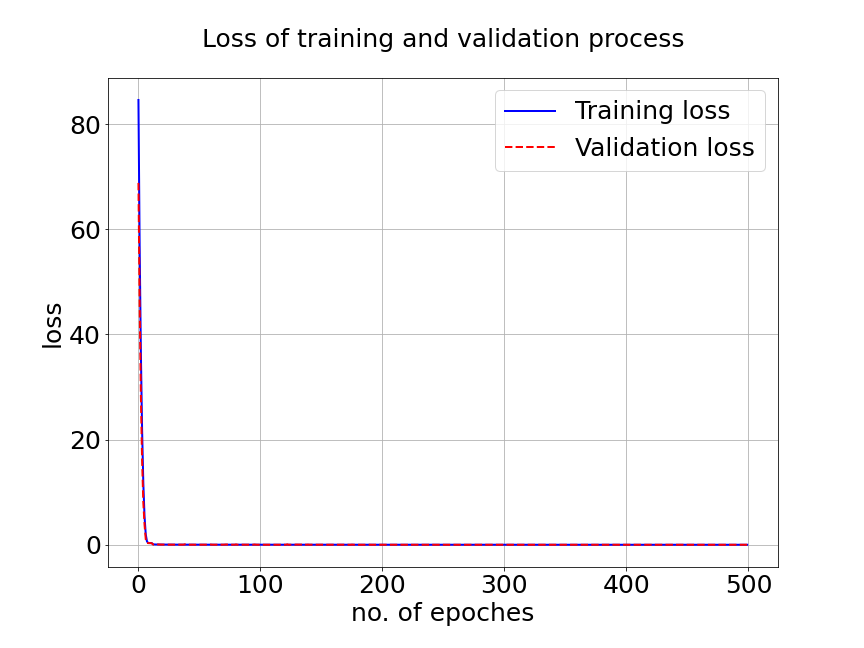}
\label{ex3_trainloss}}
\qquad 
\subfloat[b][]{
\includegraphics[width=0.45\textwidth]{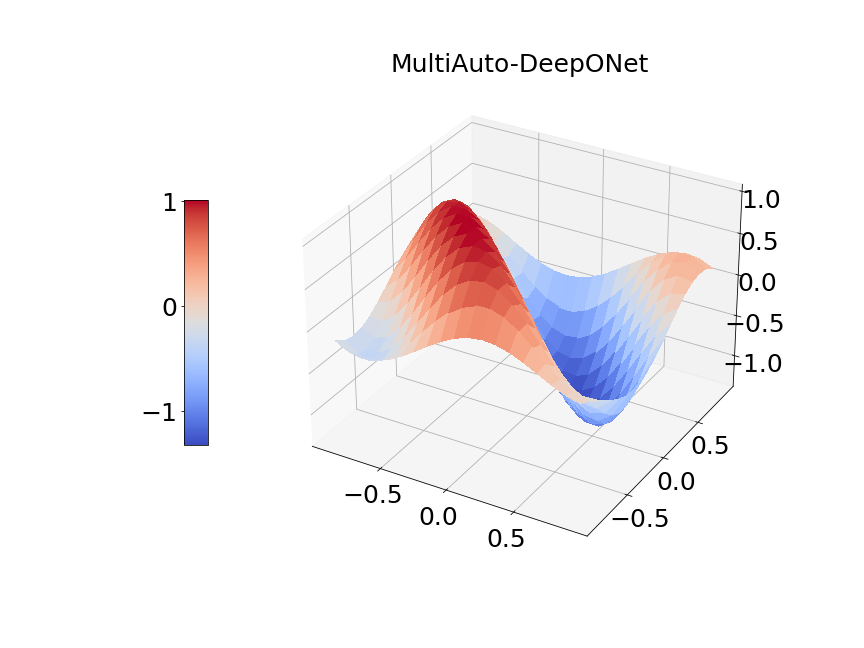}
\label{ex3_upred2}}
\qquad 
\subfloat[c][]{
\includegraphics[width=0.45\textwidth]{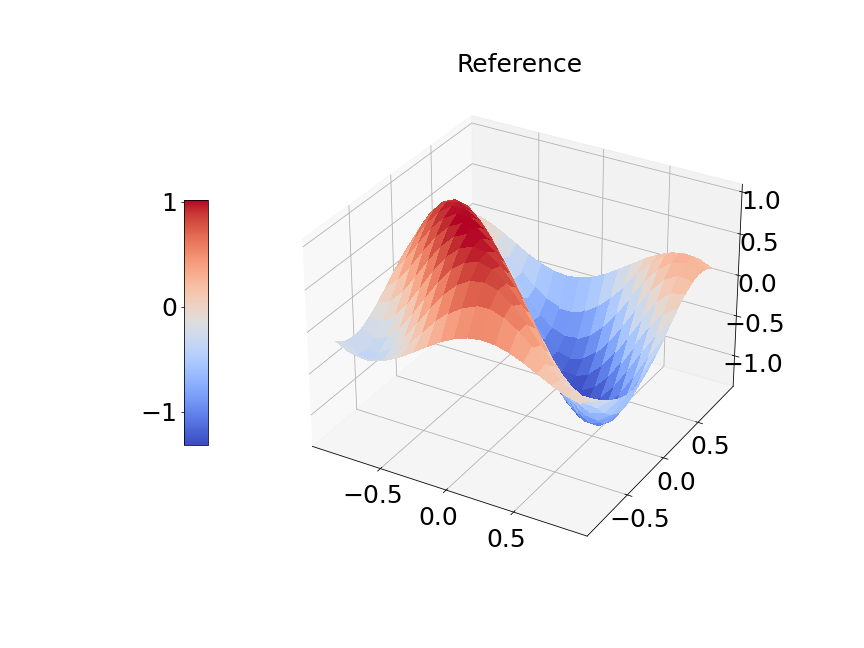}
\label{ex3_upred1}}
\qquad
\subfloat[d][]{
\includegraphics[width=0.45\textwidth]{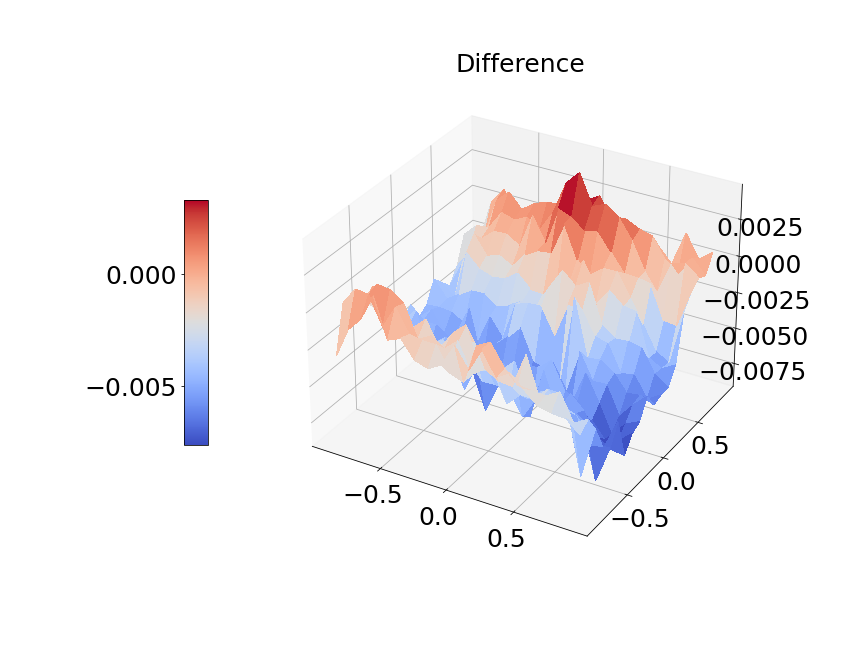}
\label{ex3_upred3}}

\caption{Two dimensional Poisson equation: (a) the training and validation loss v.s. no. of epoches. (b) MultiAuto-DeepONet model prediction of one particular sample path; (c) Reference solution of one particular sample path; (d) The difference between MultiAuto-DeepONet model prediction and reference solution.}
\label{ex3_upred}
\end{figure}

In Figure \ref{ex3_trainloss}, the blue solid line and red dashed line represent training and validation loss respectively. We can see the generalization error is very small. Figure \ref{ex3_upred2} presents the model prediction of one particular sample path for the solution $u(x, y;\omega)$. Figure \ref{ex3_upred1} is the ground true plot and Figure \ref{ex3_upred3} shows the difference plot between MultiAuto-DeepONet approximation and reference solution. We can see they match pretty well. In Table \ref{table:5}, we list the test MSE and average relative $l^2$ error of MultiAuto-DeepONet model for different number of input sensors. The number of input sample path is fixed to be $2000$ in this case. Table \ref{table:6} shows the test MSE and average $l^2$ error for different number of input sample path. The number of input sensors is fixed to be $361$. The results shown in the two tables illustrate the convergence of the model.

\begin{table}[h!]
\centering
\begin{tabular}{||c c c||} 
 \hline
 & MSE  &  Average relative $l^2$ error
 \\ [0.5ex] 
 \hline
$n_f = 81$ &  1.41E-03
&  0.0709
 \\  
 \hline
 $n_f = 225$ & 2.09E-04
&   0.0146
 \\  
 \hline
  $n_f = 361$
 & 1.43E-04
 & 0.0082
 \\  
  \hline
 $n_f = 625$
 & 1.14E-04
 & 0.0073
 \\  
  \hline
%  $n_k = 625$
%  & 
%  &  
%  \\
%   \hline
\end{tabular}
\caption{The test MSE and average $l^2$ error of MuliAuto-DeepONet model for different number of input sensors. The number of input sample path is fixed to be $2000$.}
\label{table:5}
\end{table}

\begin{table}[h!]
\centering
\begin{tabular}{||c c c||} 
 \hline
 & MSE  &  Average relative $l^2$ error
 \\ [0.5ex] 
 \hline
$n_s = 1500$ &  4.93E-04
&  0.0104
 \\  
 \hline
 $n_s = 2000$ & 1.43E-04
&   0.0082

 \\  
 \hline
 $n_s = 2500$
 & 1.22E-04
 & 0.0078
 \\  
  \hline
\end{tabular}
\caption{The test MSE and average $l^2$ error of MuliAuto-DeepONet model for different number of input sample path. The number of input sensors is fixed to be $361$.}
\label{table:6}
\end{table}

\begin{figure}[h]
\centering
\subfloat[a][]{
\includegraphics[width=0.45\textwidth]{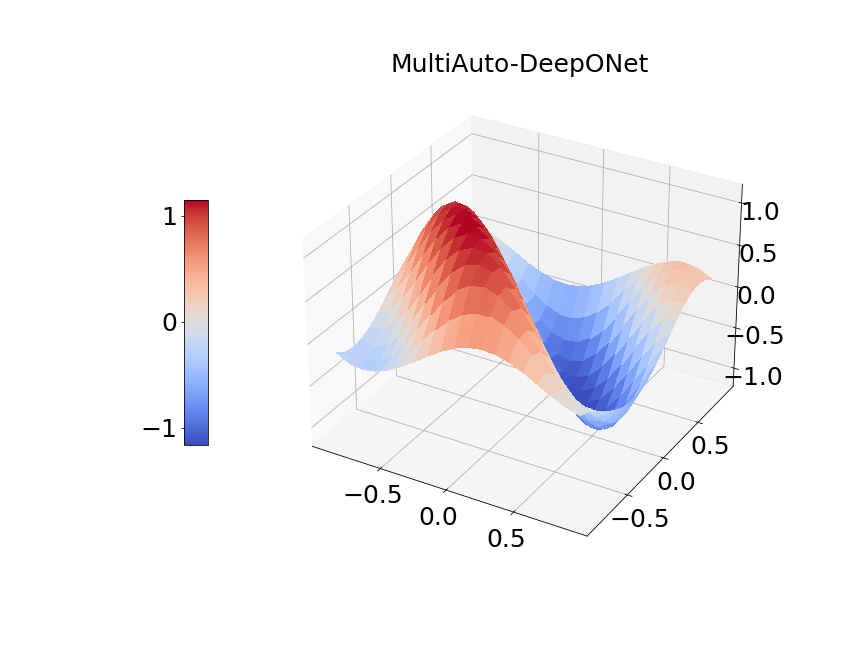}
\label{ex3_umean}}
\qquad 
\subfloat[b][]{
\includegraphics[width=0.45\textwidth]{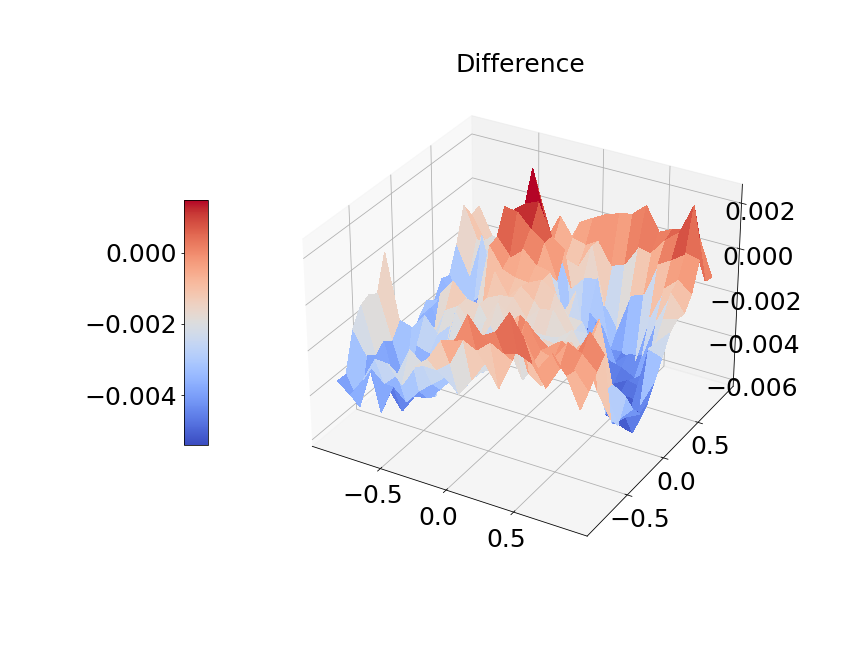}
\label{ex3_umeandiff}}
\qquad 
\subfloat[c][]{
\includegraphics[width=0.45\textwidth]{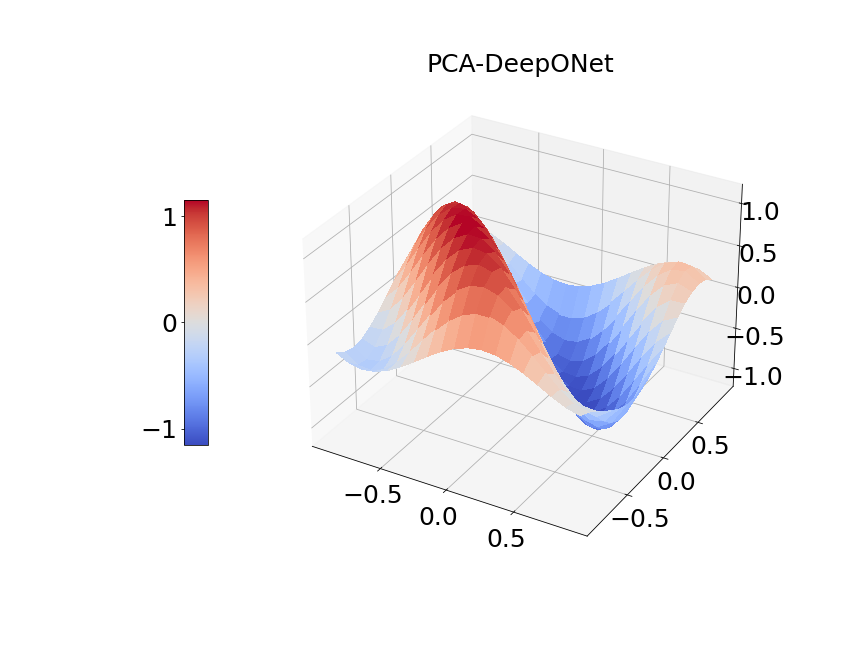}
\label{ex3_umean_PCA}}
\qquad 
\subfloat[d][]{
\includegraphics[width=0.45\textwidth]{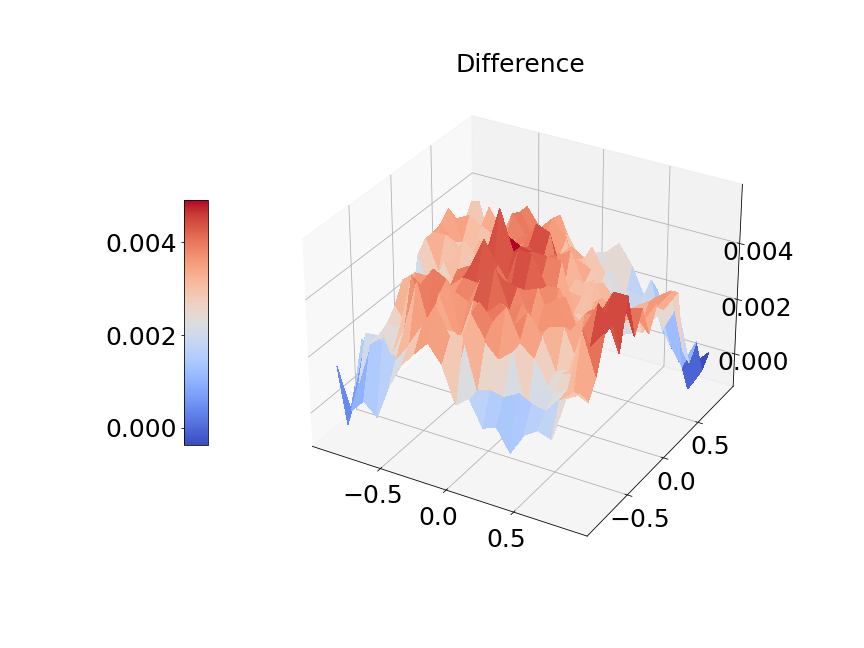}
\label{ex3_umean_PCAdiff}}
\qquad 
\subfloat[e][]{
\includegraphics[width=0.45\textwidth]{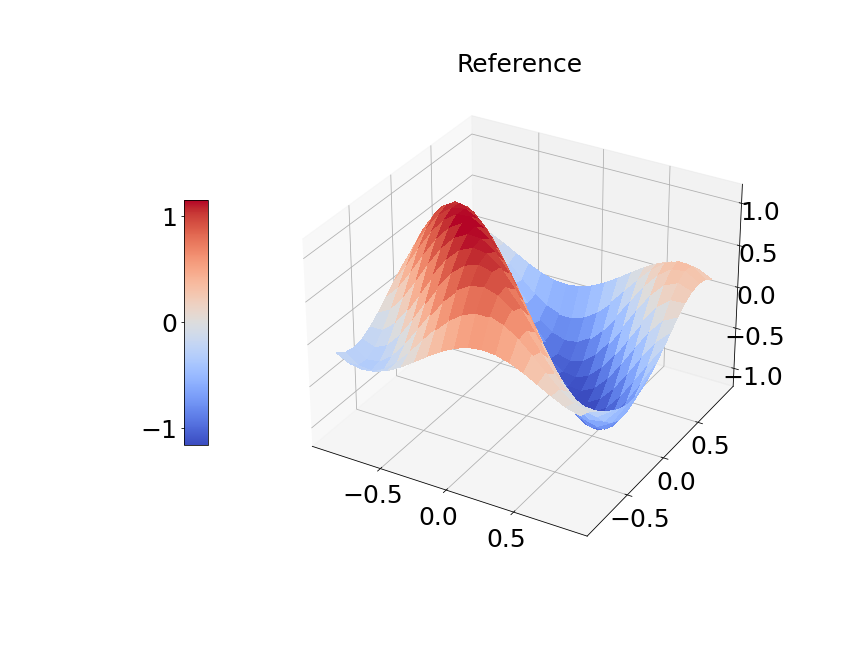}
\label{ex3_umeanre}}
\caption{Two dimensional Poisson equation: the predicted mean of different models.
(a) The predicted mean of MultiAuto-DeepONet model; (b) The difference of predicted mean between the MultiAuto-DeepONet model and reference; (c) The predicted mean of PCA based DeepONet model; (d) The difference of predicted mean between the PCA based DeepONet model and reference; (e) The mean of reference solution.
}
\label{ex3_mean}
\end{figure}

\begin{figure}[h]
\centering
\subfloat[a][]{
\includegraphics[width=0.45\textwidth]{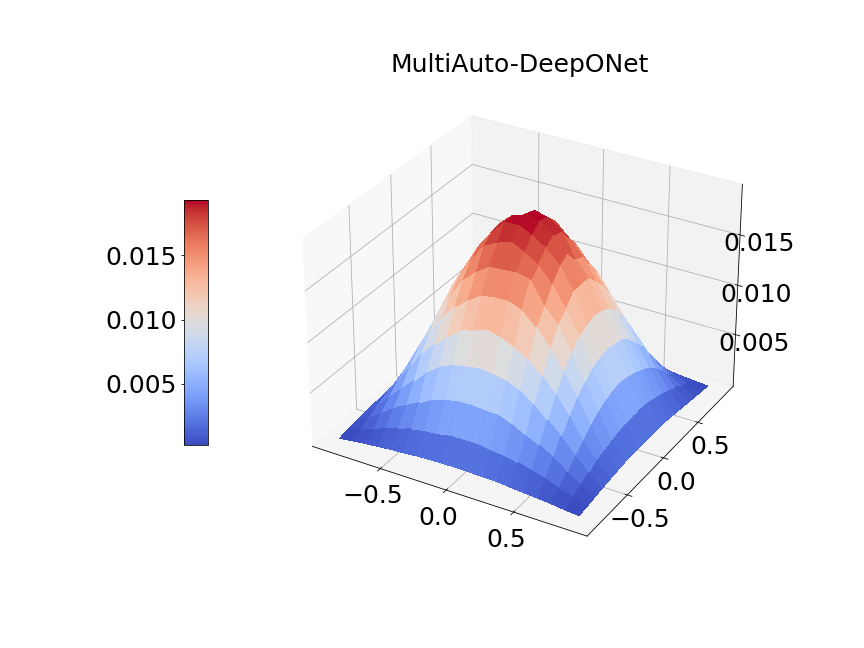}
\label{ex3_uvar1}}
\qquad 
\subfloat[b][]{
\includegraphics[width=0.45\textwidth]{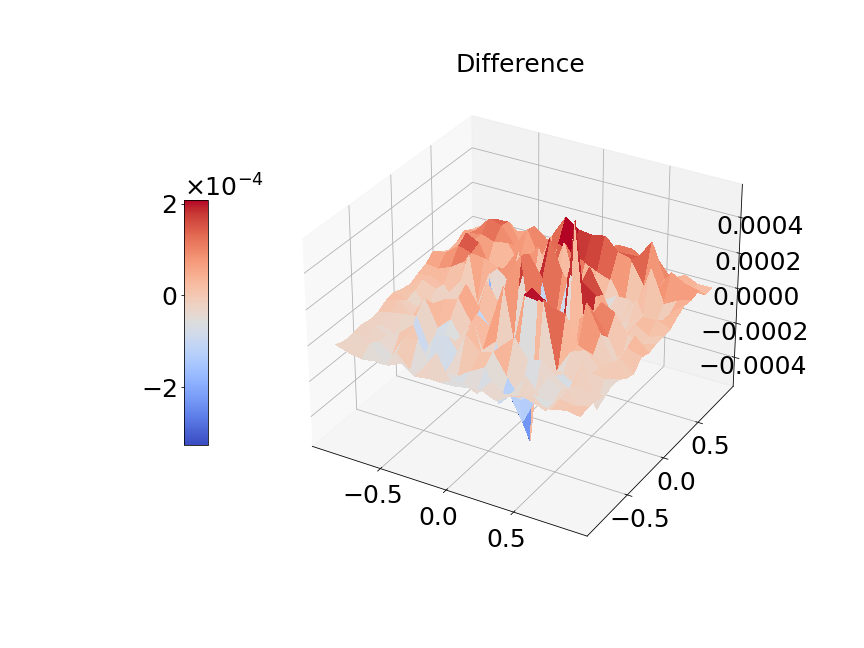}
\label{ex3_uvar3}}
\qquad 
\subfloat[c][]{
\includegraphics[width=0.45\textwidth]{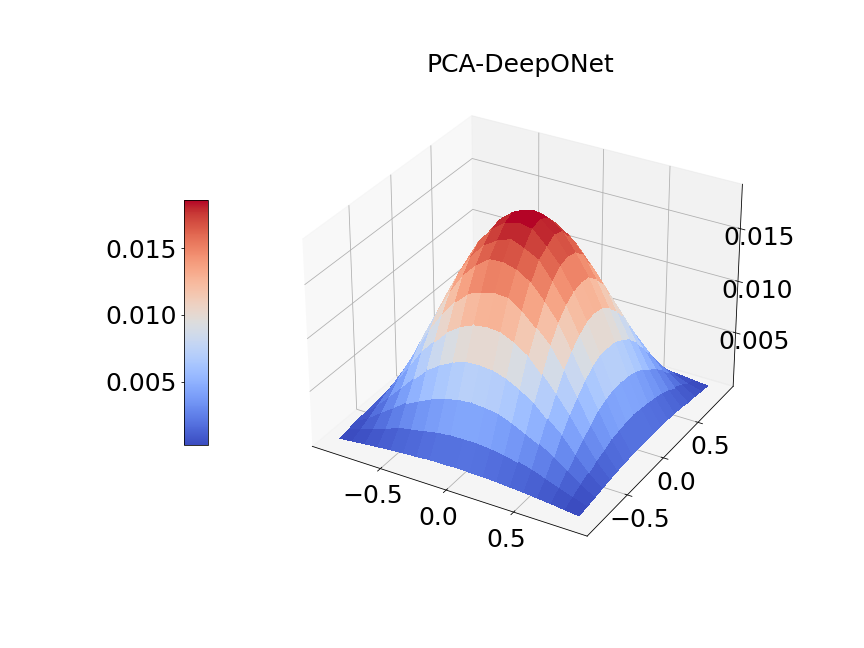}
\label{ex3_uvarPCA1}}
\qquad 
\subfloat[b][]{
\includegraphics[width=0.45\textwidth]{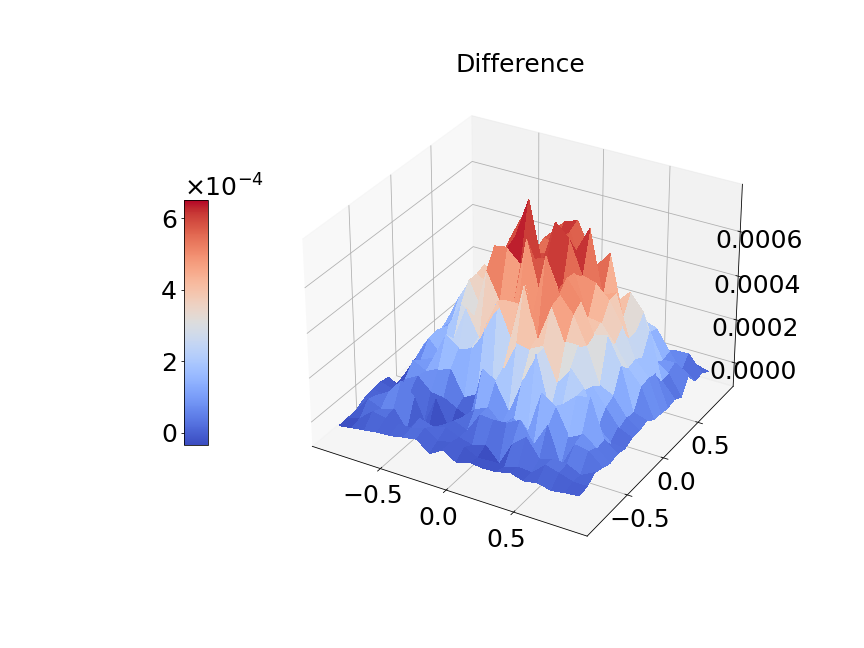}
\label{ex3_uvarPCA3}}
\qquad 
\subfloat[b][]{
\includegraphics[width=0.45\textwidth]{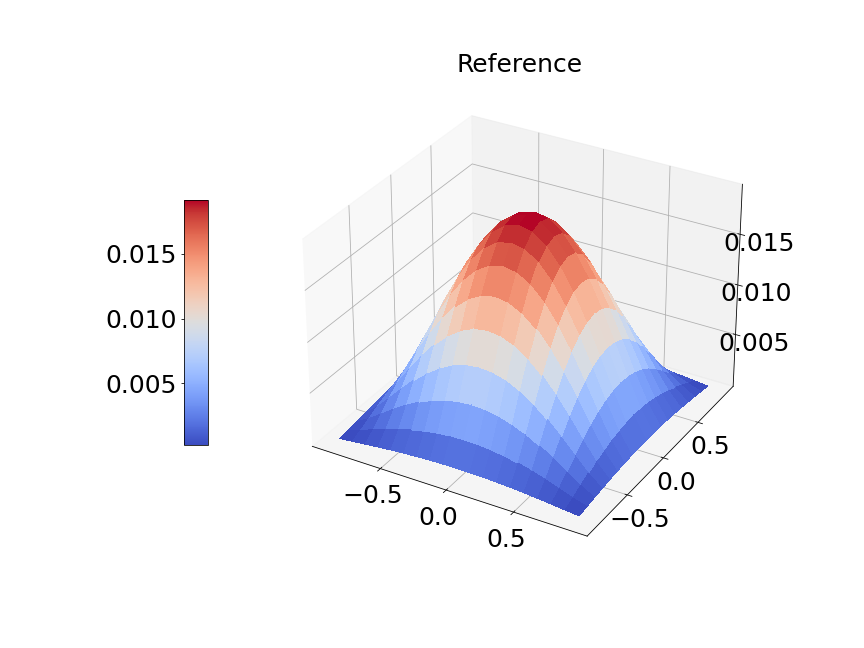}
\label{ex3_uvar2}}
\caption{Two dimensional Poisson equation: the predicted variance of different models.
(a) The predicted variance of MultiAuto-DeepONet model; (b) The difference of predicted variance between the MultiAuto-DeepONet model and reference; (c) The predicted variance of PCA based DeepONet model; (d) The difference of predicted variance between the PCA based DeepONet model and reference; (e) The variance of reference solution.}
\label{ex3_var_all}
\end{figure}

\begin{figure}[h]
\centering
\subfloat[a][]{
\includegraphics[width=0.45\textwidth]{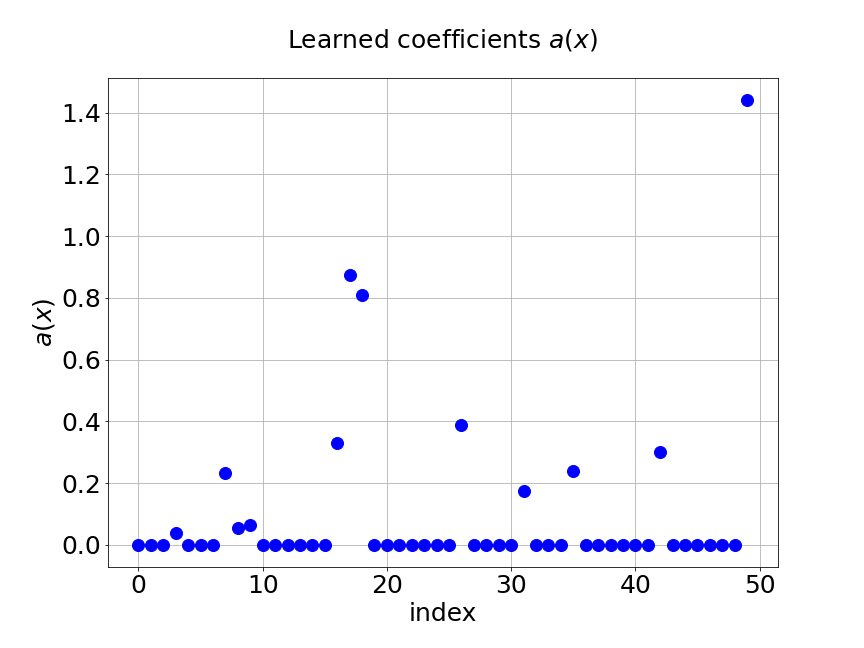}
\label{ex3_coek}}
\qquad 
\subfloat[b][]{
\includegraphics[width=0.45\textwidth]{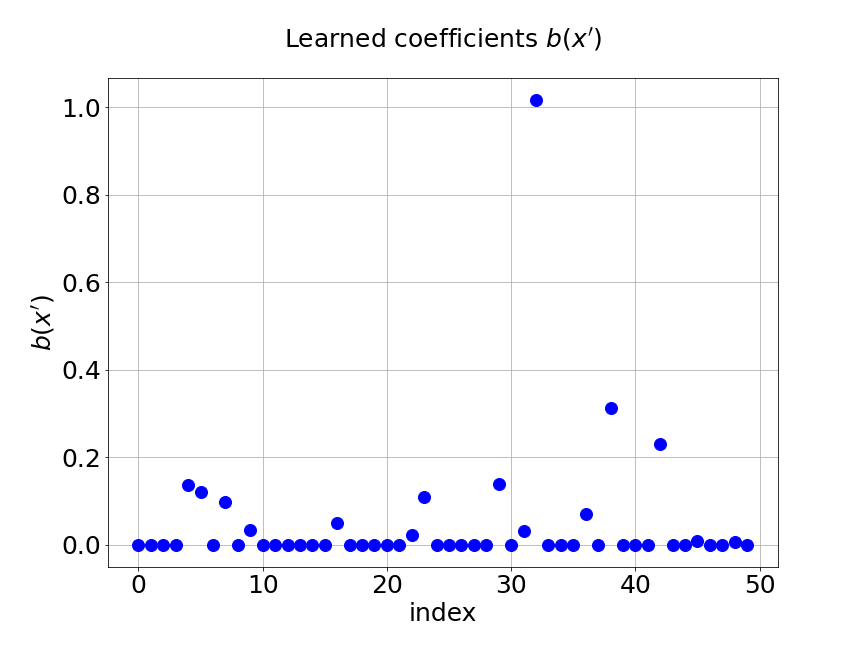}
\label{ex3_coeu}}
\qquad 
\subfloat[c][]{
\includegraphics[width=0.45\textwidth]{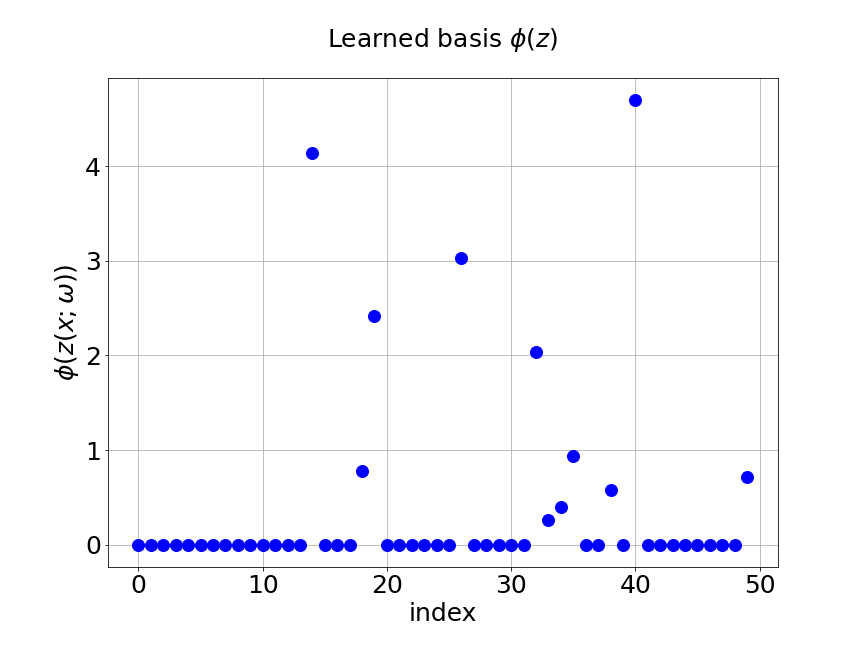}
\label{ex3_phi}}
\caption{Sparse coefficients and basis learned from MultiAuto-DeepONet: (a) The coefficients $\vec{a}(x)$ from the unsupervised trunk net; (b) The coefficients $\vec{b}(x')$ from the supervised trunk net; (c) The basis $\vec{\phi}(z)$ from the branch net.}
\label{ex3_sparse}
\end{figure}

Now, we compare our model with PCA based DeepONet for the predicted mean and variance. The number of sample trajectories of is $1800$ and the number of input sensors is $400$ for the test set. Figure \ref{ex3_umean} shows the predicted mean of our model and Figure \ref{ex3_umeandiff} shows the difference between our model and the reference. Figure \ref{ex3_umean_PCA} presents the PCA based DeepONet predicted mean and Figure \ref{ex3_umean_PCAdiff} is the corresponding difference. The mean of reference is shown in Figure \ref{ex3_umeanre}. In Figure \ref{ex3_var_all}, we present the predicted variance of our model, PCA based DeepONet model and their difference with the reference. Table \ref{table:7} shows the relative $l^2$ errors of predicted mean and variance for the two models. We can see MultiAuto-DeepONet outperforms PCA based DeepONet in this case for both mean and variance. Figure \ref{ex3_sparse} presents the coefficients and basis learned from MultiAuto-DeepONet model for one specific sample at corresponding sensor locations. We can see most of those values are zeros indicating a sparse structure.

\begin{table}[h!]
\centering
\begin{tabular}{||c c c||} 
 \hline
&  PCA based DeepONet &  MultiAuto-DeepONet 
 \\ [0.5ex] 
  \hline
 Mean error &  0.0071
 & \textbf{0.0046}
     \\
 \hline
Variance error & 0.0300
  &  \textbf{0.0243
}
 \\  
 \hline
\end{tabular}
\caption{Comparison of relative $l^2$ errors of predicted mean and variance for different models of two dimensional Poisson equation.}
\label{table:7}
\end{table}

We perform another experiment in this section to show the effectiveness of prediction basis $\phi(\vec{t})$ from the branch net in MultiAuto-DeepONet model. First we replace the branch net part, i.e. the basis $\phi(\vec{t})$ with the basis learned using polynomial chaos expansion(PCE) and fix this part during the training process of network. Then the MSE of this method is compared with our original model. Note that we choose the order of Legendre polynomials in PCE method to be $3$ considering both the accuracy and computational efficiency in this case. During the training process of MultiAuto-DeepONet model, we adjust the number of basis and coefficients learned from the branch net and trunk nets to be the same with the number of basis in PCE method. The number of input sensors is $361$ and the number of input sample path is $2000$. The test MSE of PCE method and our model are approximately $0.102\%$ and $0.047\%$ respectively. This means MultiAuto-DeepONet model can capture the basis of the SDE solution better for this forward stochastic problem.

% Table \ref{table:8} list the MSE errors of two methods.

% \begin{table}[h!]
% \centering
% \begin{tabular}{||c c c||} 
%  \hline
%  & PCE &   MultiAuto-DeepONet 
%  \\ [0.5ex] 
%   \hline
%  MSE  &  2.14e-04
%  & \textbf{4.66e-05}
%      \\
%  \hline
% \end{tabular}
% \caption{Comparison of mean square errors for different models of two dimensional Poisson equation.}
% \label{table:8}
% \end{table}

\subsection{Forward problem: Korteweg–De Vries(KdV) Equation}
In last example, we consider the forward problem of the Korteweg–De Vries (KdV) equation with time-dependent additive noise. It is a more complicated and nonlinear equation modeling waves on shallow water surfaces,
\begin{equation}
    u_t(x, t; \omega) - 6 u(x, t; \omega) u_x(x, t; \omega) + u_{xxx}(x, t; \omega) = f(t; \omega), \hspace{3mm} x \in (-\infty, \infty)
\end{equation}
with initial condition,
\begin{equation}
    u(x, 0; \omega) = -2 sech^2(x)
\end{equation}
Here, $\omega \in \Omega$ is the random space in this example. We know the analytical solution is,
\begin{equation}
    u(x, t; \omega) = W(t; \omega) - 2sech^2(x - 4t + 6 \int_0^t W(z;\omega)dz)
\end{equation}
where $W(t; \omega)$ is defined to be,
\begin{equation}
    W(t; \omega) = \int_0^t f(y;\omega)dy
\end{equation}
Here, we use KL expansion to model $f(t;\omega)$ as a Gaussian random field,
\begin{equation}
    f(t;\omega) = \sigma \sum_{i=1}^d \sqrt{\lambda_i} \phi_i(t) \omega_i
\end{equation}
where $\sigma$ is a constant and we set it to be $0.1$, $\{\lambda_i\}_{i=1}^d$ and $\{\phi_i(t)\}_{i=1}^d$ are $d$ largest eigenvalues and corresponding eigenfunctions of the covariance kernel:
\begin{equation}
    Cov(t, t') = exp(\frac{|t-t'|}{l_c})
\end{equation}
In this example, $d$ is set to be $3$ and $l_c = 0.25$. We also limit the spatial $x \in [0, 4]$ and the final time is set to be $t = 0.1$s. The input of our encoder is similar to the previous examples. The unsupervised trunk net is used to process the temporal input $t$ and the supervised trunk net is used to process the input of both spatial and temporal inputs $(x, t)$. This also illustrates the capability of our model to deal with inputs of different dimensions.

\begin{figure}[h]
\centering
\subfloat[a][]{
\includegraphics[width=0.45\textwidth]{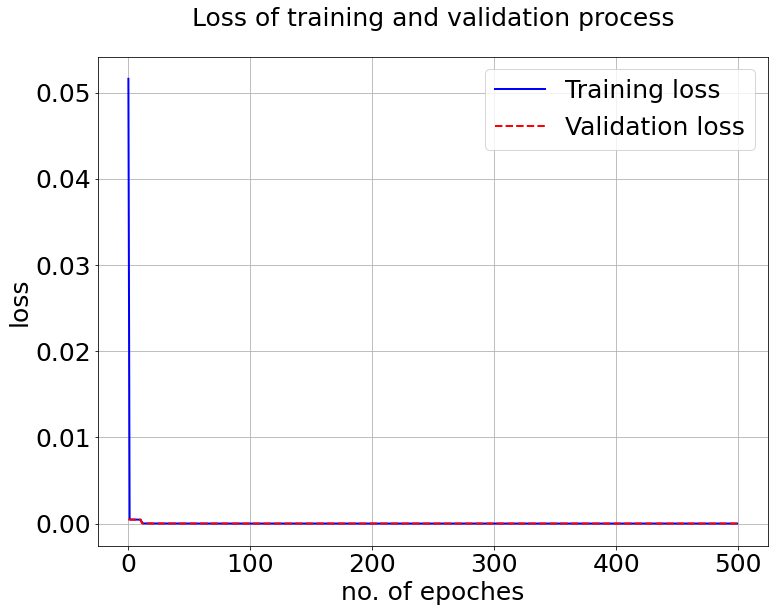}
\label{ex4_trainloss}
}
\qquad 
\subfloat[b][]{
\includegraphics[width=0.45\textwidth]{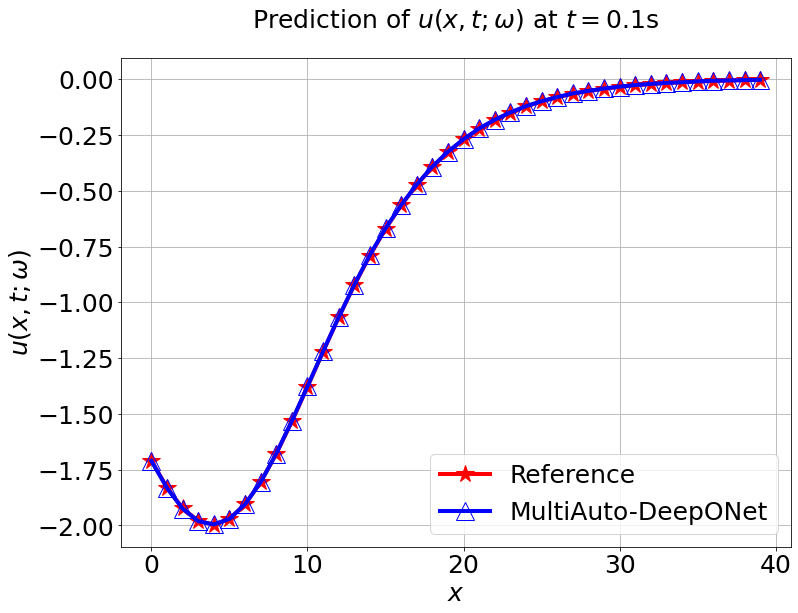}
\label{ex4_upred}}
\caption{KdV equation: (a) The training and validation loss v.s. no. of epoches. (b) The MultiAuto-DeepONet prediction of one particular sample path. The red line is the reference and the blue line is the model prediction.
}
\label{ex4_fig1}
\end{figure}

We first train the model with $4000$ sample trajectories of $f(t; \omega)$. The number of input sensor for $f(t; \omega)$ is $400$, i.e. $40 \times 10$ grid for $(x, t)$. The latent dimension for the encoder is fixed to be $4$. Other settings are the same as in two dimensional Poisson example. Figure \ref{ex4_trainloss} shows the training and validation loss. The red and blue lines match well meaning generalization error is very small. In Figure \ref{ex4_upred}, we present the model prediction and the reference solution at $t = 0.1$s of $u(x, t; \omega)$ for one sample trajectory of $f(t; \omega)$. The test MSE is about $6.34E-05$ and average relative $l^2$ error is around $0.097\%$. 

\begin{figure}[h]
\centering
\subfloat[a][]{
\includegraphics[width=0.45\textwidth]{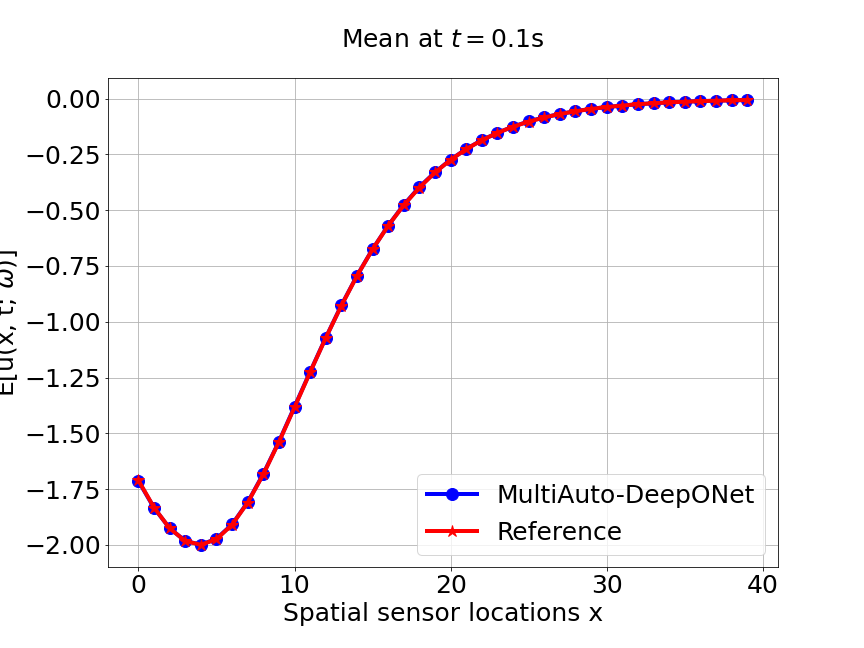}
\label{ex4_umean}}
\qquad 
\subfloat[b][]{
\includegraphics[width=0.45\textwidth]{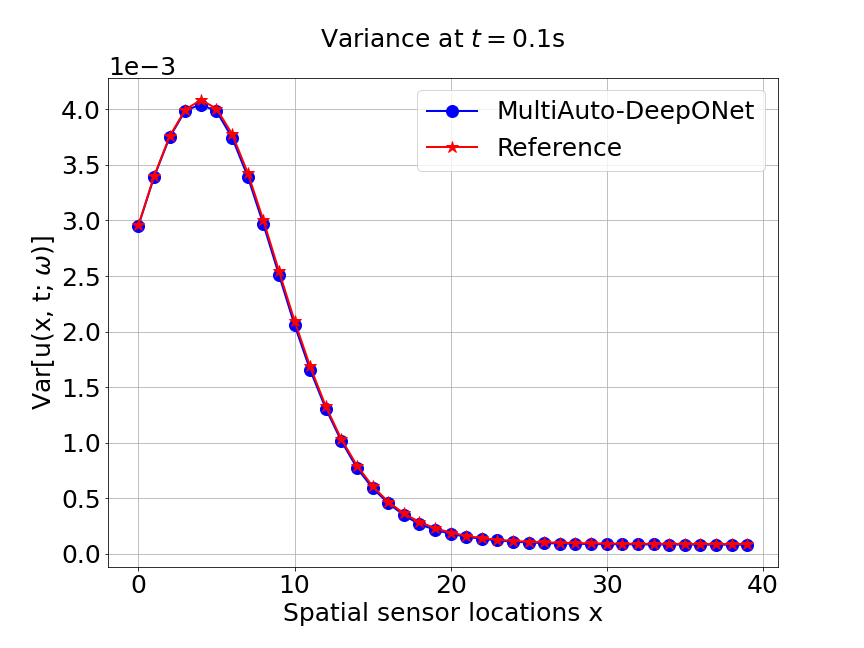}
\label{ex4_uvar}}
\qquad 
\subfloat[c][]{
\includegraphics[width=0.45\textwidth]{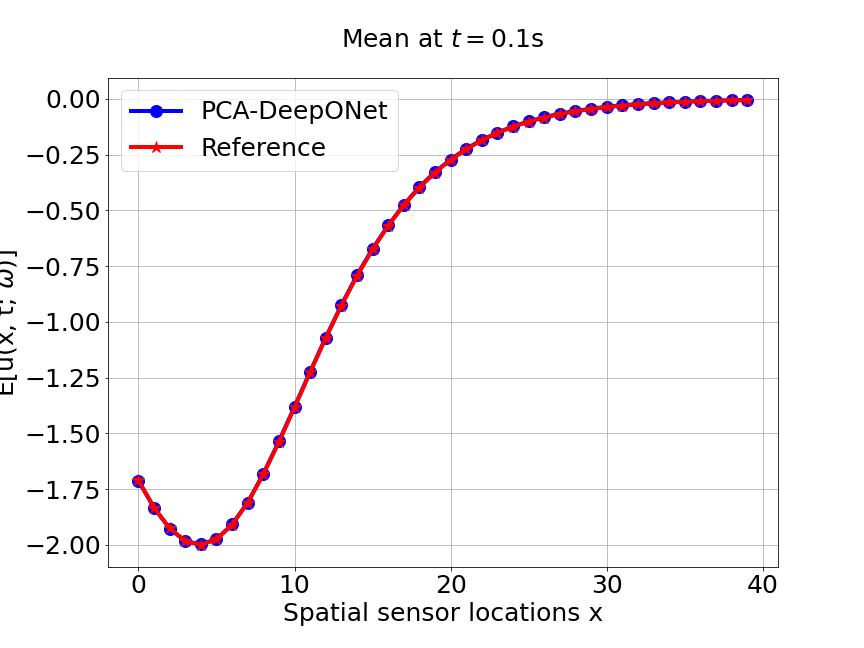}
\label{ex4_umeanPCA}}
\qquad 
\subfloat[d][]{
\includegraphics[width=0.45\textwidth]{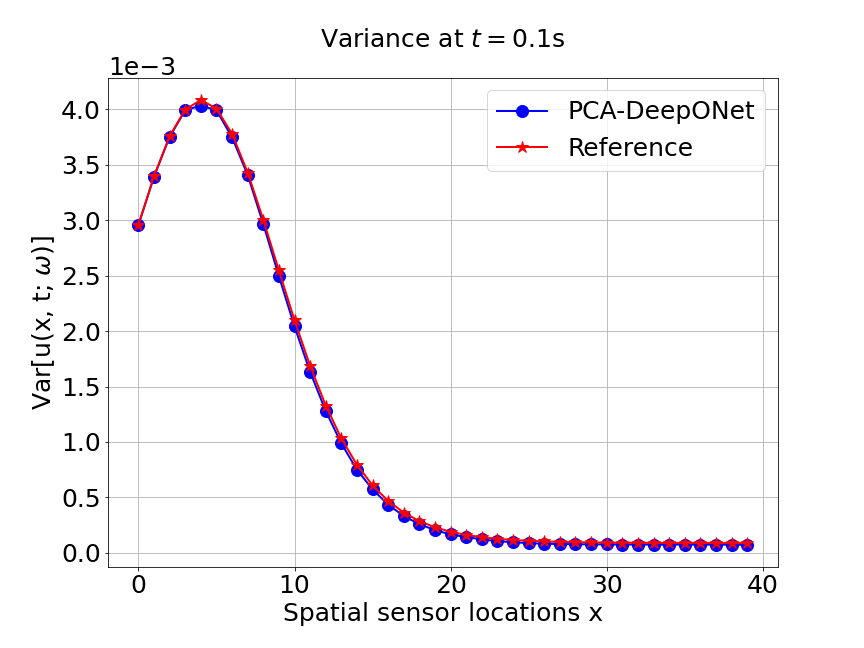}
\label{ex4_uvarPCA}}
\caption{KdV equation: the predicted mean and variance of different models at $t=0.1$s. (a) The predicted mean of $u(x, 0.1; w)$ from MultiAuto-DeepONet model and the reference solution mean; (b) The predicted variance of $u(x, 0.1; w)$ from MultiAuto-DeepONet model and the reference solution variance; (c) The predicted mean of $u(x, 0.1; w)$ from PCA based DeepONet model and the reference solution mean; (d) The predicted variance of $u(x, 0.1; w)$ from PCA based DeepONet model and the reference solution variance.
}
\label{ex4_meanvar}
\end{figure}

\begin{figure}[h]
\centering
\subfloat[a][]{
\includegraphics[width=0.45\textwidth]{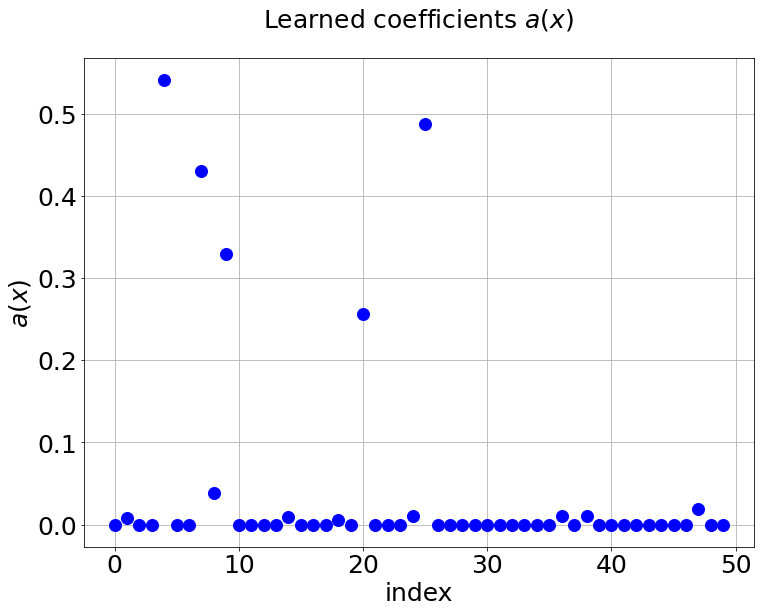}
\label{ex4_coek}}
\qquad 
\subfloat[b][]{
\includegraphics[width=0.45\textwidth]{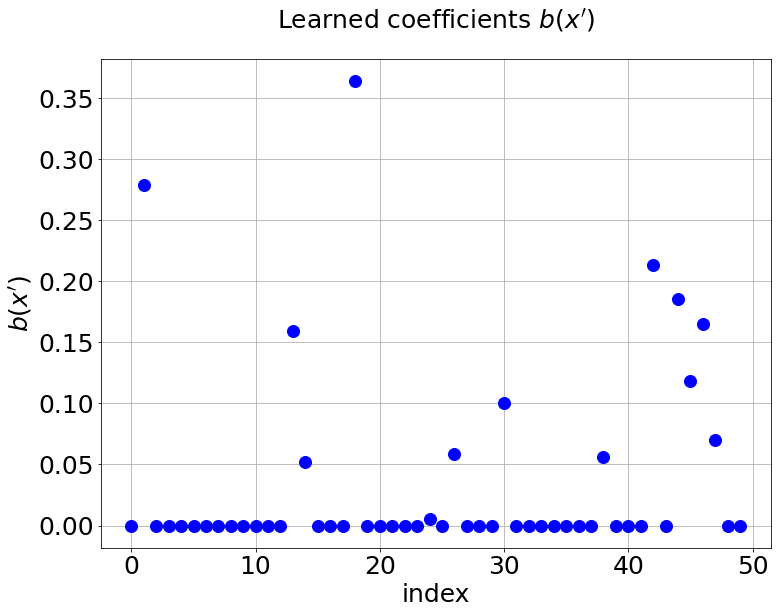}
\label{ex4_coeu}}
\qquad 
\subfloat[c][]{
\includegraphics[width=0.45\textwidth]{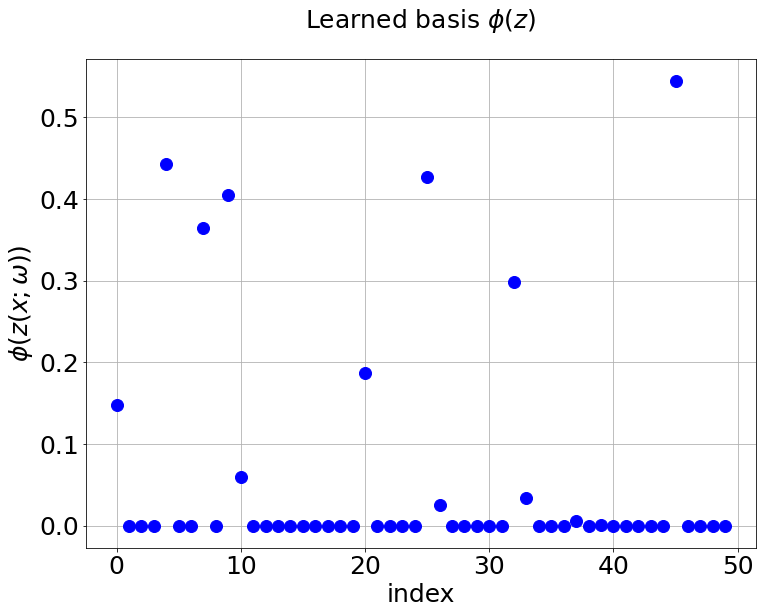}
\label{ex4_phi}}
\caption{Sparse coefficients and basis learned from MultiAuto-DeepONet: (a) The coefficients $\vec{a}(x)$ from the unsupervised trunk net; (b) The coefficients $\vec{b}(x')$ from the supervised trunk net; (c) The basis $\vec{\phi}(z)$ from the branch net.}
\label{ex4_sparse}
\end{figure}

Now, we compare our model with PCA based DeepONet for the predicted mean and variance. The number of sample trajectories of is $2000$ for the test set and the number of input sensor is the same as training data set. Figure \ref{ex4_umean} shows the predicted mean of our model and Figure \ref{ex4_umeanPCA} shows the predicted mean of PCA based DeepONet model at final time $t=0.1$s. In Figures \ref{ex4_uvar} and \ref{ex4_uvarPCA}, the predicted variance of our model and PCA based DeepONet model are presented. Table \ref{table:8} shows the relative $l^2$ errors of predicted mean and variance for the two models. We can conclude that our model achieves better accuracy in terms of predicted mean and variance in this example. Again, Figure \ref{ex4_sparse} presents the sparse coefficients and basis learned from MultiAuto-DeepONet model for one specific sample at corresponding sensor locations.

\begin{table}[h!]
\centering
\begin{tabular}{||c c c||} 
 \hline
&  PCA based DeepONet &  MultiAuto-DeepONet 
 \\ [0.5ex] 
  \hline
 Mean error & 0.00166 & \textbf{0.00107}
     \\
 \hline
Variance error & 0.0153 & \textbf{0.0099}
 \\  
 \hline
\end{tabular}
\caption{Comparison of relative $l^2$ errors of predicted mean and variance for different models of KDV equation.}
\label{table:8}
\end{table}

%%%%%%%%%%%%%%%%%%%%%%%%%%%%%%%%%%%%%%%%%%%%%%%%%%%%%%%%%%%%%%%%%%%%%%%%%%%%%%%%%%%%%%%%%%%%%%%%%%%%%%%%%%%%%%%%%%%%%%

\section{Conclusion}\label{Summary}
In this paper, we proposes the MultiAuto-DeepONet model to solve forward and inverse problem of stochastic differential equations. In particular, we design a new network structure which equips DeepONet with convolutional autoencoder. This design solves the SDE problems with DeepONet under high-dimensional settings. It also enables us to perform nonlinear dimension reduction and operator learning simultaneously. We also show the hidden features discovered by encoder can be used to generate samples of the input function. Furthermore, we perform uncertainty quantification and compare our model with PCA based DeepONet in our numerical experiments. The first numerical example is a forward problem of a simple stochastic ODE, the goal is to demonstrate the workflow of our method in detail. We also made a comparison with the original DeepONet and PCA based DeepONet in this example. The next two examples involve one and two dimensional stochastic Poisson equation respectively. PCA based DeepONet is compared with our model to show the effectiveness of MultiAuto-DeepONet in high-dimensional settings. In the two dimensional Poisson equation case, we also
conduct an experiment of using polynomial chaos expansion to learn the basis for SDE solutions and compare it with our model to show the effectiveness of our method to learn the nonlinear basis. Finally, we present the forward problem of KdV equation to show our model can deal with inputs from different domains or with different scales. We add $L_1$ regularization in our network to introduce sparsity in this paper. The future work includes investigating the impact of different regularization terms, for example, $L_2$ and physics-informed regularization to the performance of our model. We will also present our works on learning more complex, especially higher dimensional stochastic operators in the future.

%%Vancouver style references.
\bibliographystyle{abbrv}
\bibliography{mybib}

\begin{thebibliography}{10}

\bibitem{abdar2021review}
M.~Abdar, F.~Pourpanah, S.~Hussain, D.~Rezazadegan, L.~Liu, M.~Ghavamzadeh,
  P.~Fieguth, X.~Cao, A.~Khosravi, U.~R. Acharya, et~al.
\newblock A review of uncertainty quantification in deep learning: Techniques,
  applications and challenges.
\newblock {\em Information Fusion}, 2021.

\bibitem{bellman1966dynamic}
R.~Bellman.
\newblock Dynamic programming.
\newblock {\em Science}, 153(3731):34--37, 1966.

\bibitem{cai2021deepm}
S.~Cai, Z.~Wang, L.~Lu, T.~A. Zaki, and G.~E. Karniadakis.
\newblock Deepm\&mnet: Inferring the electroconvection multiphysics fields
  based on operator approximation by neural networks.
\newblock {\em Journal of Computational Physics}, 436:110296, 2021.

\bibitem{chen2018neural}
R.~T. Chen, Y.~Rubanova, J.~Bettencourt, and D.~Duvenaud.
\newblock Neural ordinary differential equations.
\newblock {\em arXiv preprint arXiv:1806.07366}, 2018.

\bibitem{chen1993approximations}
T.~Chen and H.~Chen.
\newblock Approximations of continuous functionals by neural networks with
  application to dynamic systems.
\newblock {\em IEEE Transactions on Neural networks}, 4(6):910--918, 1993.

\bibitem{chen1995universal}
T.~Chen and H.~Chen.
\newblock Universal approximation to nonlinear operators by neural networks
  with arbitrary activation functions and its application to dynamical systems.
\newblock {\em IEEE Transactions on Neural Networks}, 6(4):911--917, 1995.

\bibitem{chen2019numerical}
T.~Chen, B.~Rozovskii, and C.-W. Shu.
\newblock Numerical solutions of stochastic pdes driven by arbitrary type of
  noise.
\newblock {\em Stochastics and Partial Differential Equations: Analysis and
  Computations}, 7(1):1--39, 2019.

\bibitem{chen2020physics}
Y.~Chen, L.~Lu, G.~E. Karniadakis, and L.~Dal~Negro.
\newblock Physics-informed neural networks for inverse problems in nano-optics
  and metamaterials.
\newblock {\em Optics express}, 28(8):11618--11633, 2020.

\bibitem{del2021learning}
J.~del {\'A}guila~Ferrandis, M.~S. Triantafyllou, C.~Chryssostomidis, and G.~E.
  Karniadakis.
\newblock Learning functionals via lstm neural networks for predicting vessel
  dynamics in extreme sea states.
\newblock {\em Proceedings of the Royal Society A}, 477(2245):20190897, 2021.

\bibitem{desmond2019symplectic}
Y.~Desmond~Zhong, B.~Dey, and A.~Chakraborty.
\newblock Symplectic ode-net: Learning hamiltonian dynamics with control.
\newblock {\em arXiv e-prints}, pages arXiv--1909, 2019.

\bibitem{di2009malliavin}
G.~Di~Nunno, B.~K. {\O}ksendal, and F.~Proske.
\newblock {\em Malliavin calculus for L{\'e}vy processes with applications to
  finance}, volume~2.
\newblock Springer, 2009.

\bibitem{dong2021local}
S.~Dong and Z.~Li.
\newblock Local extreme learning machines and domain decomposition for solving
  linear and nonlinear partial differential equations.
\newblock {\em Computer Methods in Applied Mechanics and Engineering},
  387:114129, 2021.

\bibitem{fang2016fine}
J.~Fang, Y.~Zhou, Y.~Yu, and S.~Du.
\newblock Fine-grained vehicle model recognition using a coarse-to-fine
  convolutional neural network architecture.
\newblock {\em IEEE Transactions on Intelligent Transportation Systems},
  18(7):1782--1792, 2016.

\bibitem{ghanem1990polynomial}
R.~Ghanem and P.~D. Spanos.
\newblock Polynomial chaos in stochastic finite elements.
\newblock 1990.

\bibitem{goswami2022physics}
S.~Goswami, M.~Yin, Y.~Yu, and G.~E. Karniadakis.
\newblock A physics-informed variational deeponet for predicting crack path in
  quasi-brittle materials.
\newblock {\em Computer Methods in Applied Mechanics and Engineering},
  391:114587, 2022.

\bibitem{graepel2003solving}
T.~Graepel.
\newblock Solving noisy linear operator equations by gaussian processes:
  Application to ordinary and partial differential equations.
\newblock In {\em ICML}, volume~3, pages 234--241, 2003.

\bibitem{hanin2019universal}
B.~Hanin.
\newblock Universal function approximation by deep neural nets with bounded
  width and relu activations.
\newblock {\em Mathematics}, 7(10):992, 2019.

\bibitem{hasan2020identifying}
A.~Hasan, J.~M. Pereira, S.~Farsiu, and V.~Tarokh.
\newblock Identifying latent stochastic differential equations with variational
  auto-encoders.
\newblock {\em arXiv preprint arXiv:2007.06075}, 2020.

\bibitem{hebb2005organization}
D.~O. Hebb.
\newblock {\em The organization of behavior: A neuropsychological theory}.
\newblock Psychology Press, 2005.

\bibitem{holden1996stochastic}
H.~Holden, B.~{\O}ksendal, J.~Ub{\o}e, and T.~Zhang.
\newblock Stochastic partial differential equations.
\newblock In {\em Stochastic partial differential equations}, pages 141--191.
  Springer, 1996.

\bibitem{hornik1989multilayer}
K.~Hornik, M.~Stinchcombe, and H.~White.
\newblock Multilayer feedforward networks are universal approximators.
\newblock {\em Neural networks}, 2(5):359--366, 1989.

\bibitem{hullermeier2019aleatoric}
E.~H{\"u}llermeier and W.~Waegeman.
\newblock Aleatoric and epistemic uncertainty in machine learning: A tutorial
  introduction.
\newblock 2019.

\bibitem{jia2019neural}
J.~Jia and A.~R. Benson.
\newblock Neural jump stochastic differential equations.
\newblock {\em arXiv preprint arXiv:1905.10403}, 2019.

\bibitem{jin2021efficient}
H.~Jin, J.~Yang, and S.~Zhang.
\newblock Efficient action recognition with introducing r (2+ 1) d convolution
  to improved transformer.
\newblock In {\em 2021 4th International Conference on Information
  Communication and Signal Processing (ICICSP)}, pages 379--383. IEEE, 2021.

\bibitem{kabir2020spinalnet}
H.~Kabir, M.~Abdar, S.~M.~J. Jalali, A.~Khosravi, A.~F. Atiya, S.~Nahavandi,
  and D.~Srinivasan.
\newblock Spinalnet: Deep neural network with gradual input.
\newblock {\em arXiv preprint arXiv:2007.03347}, 2020.

\bibitem{kendall2015bayesian}
A.~Kendall, V.~Badrinarayanan, and R.~Cipolla.
\newblock Bayesian segnet: Model uncertainty in deep convolutional
  encoder-decoder architectures for scene understanding.
\newblock {\em arXiv preprint arXiv:1511.02680}, 2015.

\bibitem{khoo2021solving}
Y.~Khoo, J.~Lu, and L.~Ying.
\newblock Solving parametric pde problems with artificial neural networks.
\newblock {\em European Journal of Applied Mathematics}, 32(3):421--435, 2021.

\bibitem{kingma2013auto}
D.~P. Kingma and M.~Welling.
\newblock Auto-encoding variational bayes.
\newblock {\em arXiv preprint arXiv:1312.6114}, 2013.

\bibitem{lagaris1998artificial}
I.~E. Lagaris, A.~Likas, and D.~I. Fotiadis.
\newblock Artificial neural networks for solving ordinary and partial
  differential equations.
\newblock {\em IEEE transactions on neural networks}, 9(5):987--1000, 1998.

\bibitem{lagaris2000neural}
I.~E. Lagaris, A.~C. Likas, and D.~G. Papageorgiou.
\newblock Neural-network methods for boundary value problems with irregular
  boundaries.
\newblock {\em IEEE Transactions on Neural Networks}, 11(5):1041--1049, 2000.

\bibitem{DONet}
L.~Lu, P.~Jin, G.~Pang, Z.~Zhang, and G.~E. Karniadakis.
\newblock Learning nonlinear operators via deeponet based on the universal
  approximation theorem of operators.
\newblock {\em Nature Machine Intelligence}, 3(3):218--229, 2021.

\bibitem{lu2019dying}
L.~Lu, Y.~Shin, Y.~Su, and G.~E. Karniadakis.
\newblock Dying relu and initialization: Theory and numerical examples.
\newblock {\em arXiv preprint arXiv:1903.06733}, 2019.

\bibitem{mao2021deepm}
Z.~Mao, L.~Lu, O.~Marxen, T.~A. Zaki, and G.~E. Karniadakis.
\newblock Deepm\&mnet for hypersonics: Predicting the coupled flow and
  finite-rate chemistry behind a normal shock using neural-network
  approximation of operators.
\newblock {\em Journal of Computational Physics}, 447:110698, 2021.

\bibitem{menter1994two}
F.~R. Menter.
\newblock Two-equation eddy-viscosity turbulence models for engineering
  applications.
\newblock {\em AIAA journal}, 32(8):1598--1605, 1994.

\bibitem{milstein2009solving}
G.~Milstein and M.~Tretyakov.
\newblock Solving parabolic stochastic partial differential equations via
  averaging over characteristics.
\newblock {\em Mathematics of computation}, 78(268):2075--2106, 2009.

\bibitem{mitzenmacher2017probability}
M.~Mitzenmacher and E.~Upfal.
\newblock {\em Probability and computing: Randomization and probabilistic
  techniques in algorithms and data analysis}.
\newblock Cambridge university press, 2017.

\bibitem{mukhoti2018evaluating}
J.~Mukhoti and Y.~Gal.
\newblock Evaluating bayesian deep learning methods for semantic segmentation.
\newblock {\em arXiv preprint arXiv:1811.12709}, 2018.

\bibitem{nabian2018deep}
M.~A. Nabian and H.~Meidani.
\newblock A deep neural network surrogate for high-dimensional random partial
  differential equations.
\newblock {\em arXiv preprint arXiv:1806.02957}, 2018.

\bibitem{oja1982simplified}
E.~Oja.
\newblock Simplified neuron model as a principal component analyzer.
\newblock {\em Journal of mathematical biology}, 15(3):267--273, 1982.

\bibitem{pang2019neural}
G.~Pang, L.~Yang, and G.~E. Karniadakis.
\newblock Neural-net-induced gaussian process regression for function
  approximation and pde solution.
\newblock {\em Journal of Computational Physics}, 384:270--288, 2019.

\bibitem{qin2021deep}
T.~Qin, Z.~Chen, J.~D. Jakeman, and D.~Xiu.
\newblock Deep learning of parameterized equations with applications to
  uncertainty quantification.
\newblock {\em International Journal for Uncertainty Quantification}, 11(2),
  2021.

\bibitem{qu2022learning}
J.~Qu, W.~Cai, and Y.~Zhao.
\newblock Learning time-dependent pdes with a linear and nonlinear separate
  convolutional neural network.
\newblock {\em Journal of Computational Physics}, page 110928, 2022.

\bibitem{qu2021features}
L.~Qu, J.~Lyu, W.~Li, D.~Ma, and H.~Fan.
\newblock Features injected recurrent neural networks for short-term traffic
  speed prediction.
\newblock {\em Neurocomputing}, 451:290--304, 2021.

\bibitem{raissi2018hidden}
M.~Raissi and G.~E. Karniadakis.
\newblock Hidden physics models: Machine learning of nonlinear partial
  differential equations.
\newblock {\em Journal of Computational Physics}, 357:125--141, 2018.

\bibitem{raissi2017machine}
M.~Raissi, P.~Perdikaris, and G.~E. Karniadakis.
\newblock Machine learning of linear differential equations using gaussian
  processes.
\newblock {\em Journal of Computational Physics}, 348:683--693, 2017.

\bibitem{NumGP}
M.~Raissi, P.~Perdikaris, and G.~E. Karniadakis.
\newblock Numerical gaussian processes for time-dependent and nonlinear partial
  differential equations.
\newblock {\em SIAM Journal on Scientific Computing}, 40(1):A172--A198, 2018.

\bibitem{PINN}
M.~Raissi, P.~Perdikaris, and G.~E. Karniadakis.
\newblock Physics-informed neural networks: A deep learning framework for
  solving forward and inverse problems involving nonlinear partial differential
  equations.
\newblock {\em Journal of Computational Physics}, 378:686--707, 2019.

\bibitem{rudy2017data}
S.~H. Rudy, S.~L. Brunton, J.~L. Proctor, and J.~N. Kutz.
\newblock Data-driven discovery of partial differential equations.
\newblock {\em Science Advances}, 3(4):e1602614, 2017.

\bibitem{rumelhart1985learning}
D.~E. Rumelhart, G.~E. Hinton, and R.~J. Williams.
\newblock Learning internal representations by error propagation.
\newblock Technical report, California Univ San Diego La Jolla Inst for
  Cognitive Science, 1985.

\bibitem{sarkka2011linear}
S.~S{\"a}rkk{\"a}.
\newblock Linear operators and stochastic partial differential equations in
  gaussian process regression.
\newblock In {\em International Conference on Artificial Neural Networks},
  pages 151--158. Springer, 2011.

\bibitem{wang2021bayesian}
Y.~Wang, W.~Deng, and G.~Lin.
\newblock Bayesian sparse learning with preconditioned stochastic gradient mcmc
  and its applications.
\newblock {\em Journal of Computational Physics}, 432:110134, 2021.

\bibitem{weinan2011principles}
E.~Weinan.
\newblock {\em Principles of multiscale modeling}.
\newblock Cambridge University Press, 2011.

\bibitem{winovich2019convpde}
N.~Winovich, K.~Ramani, and G.~Lin.
\newblock Convpde-uq: Convolutional neural networks with quantified uncertainty
  for heterogeneous elliptic partial differential equations on varied domains.
\newblock {\em Journal of Computational Physics}, 394:263--279, 2019.

\bibitem{xiu2002wiener}
D.~Xiu and G.~E. Karniadakis.
\newblock The wiener--askey polynomial chaos for stochastic differential
  equations.
\newblock {\em SIAM journal on scientific computing}, 24(2):619--644, 2002.

\bibitem{yang2016enhancing}
X.~Yang, H.~Lei, N.~A. Baker, and G.~Lin.
\newblock Enhancing sparsity of hermite polynomial expansions by iterative
  rotations.
\newblock {\em Journal of Computational Physics}, 307:94--109, 2016.

\bibitem{zhang2019quantifying}
D.~Zhang, L.~Lu, L.~Guo, and G.~E. Karniadakis.
\newblock Quantifying total uncertainty in physics-informed neural networks for
  solving forward and inverse stochastic problems.
\newblock {\em Journal of Computational Physics}, 397:108850, 2019.

\bibitem{zhang2021deep}
Y.~Zhang, Y.~Chen, and C.~Gao.
\newblock Deep unsupervised multi-modal fusion network for detecting driver
  distraction.
\newblock {\em Neurocomputing}, 421:26--38, 2021.

\bibitem{zhang2021hidden}
Y.~Zhang, X.~Zhu, and J.~Gao.
\newblock Hidden physics model for parameter estimation of elastic wave
  equations.
\newblock {\em Computer Methods in Applied Mechanics and Engineering},
  381:113814, 2021.

\bibitem{zhou2021solving}
Z.~Zhou and Z.~Yan.
\newblock Solving forward and inverse problems of the logarithmic nonlinear
  schr{\"o}dinger equation with pt-symmetric harmonic potential via deep
  learning.
\newblock {\em Physics Letters A}, 387:127010, 2021.

\bibitem{zhu2018bayesian}
Y.~Zhu and N.~Zabaras.
\newblock Bayesian deep convolutional encoder--decoder networks for surrogate
  modeling and uncertainty quantification.
\newblock {\em Journal of Computational Physics}, 366:415--447, 2018.

\bibitem{zhu2019physics}
Y.~Zhu, N.~Zabaras, P.-S. Koutsourelakis, and P.~Perdikaris.
\newblock Physics-constrained deep learning for high-dimensional surrogate
  modeling and uncertainty quantification without labeled data.
\newblock {\em Journal of Computational Physics}, 394:56--81, 2019.

\end{thebibliography}
\nocite{ghanem1990polynomial,xiu2002wiener,sarkka2011linear,lagaris1998artificial,nabian2018deep,zhu2018bayesian,kingma2013auto,kendall2015bayesian,weinan2011principles,del2021learning,qin2021deep,chen2018neural,jia2019neural,desmond2019symplectic,winovich2019convpde,zhu2019physics,rudy2017data,chen2020physics,hornik1989multilayer,chen1993approximations,lu2019dying,mitzenmacher2017probability,cai2021deepm,hanin2019universal,mao2021deepm,chen2019numerical,jin2021efficient,zhang2019quantifying,menter1994two,dong2021local,wang2021bayesian,fang2016fine,goswami2022physics,zhang2021hidden,qu2021features,zhou2021solving,qu2022learning,zhang2021deep,hasan2020identifying,yang2016enhancing}
    
\end{document}